\documentclass[a4paper, twoside, leqno]{article}
\usepackage{PRIMEarxiv}
\renewcommand{\normalsize}{\fontsize{12}{12}\selectfont} 
\usepackage[skip=2pt,font=small]{caption}
\DeclareCaptionFont{Cfont}{\fontsize{12}{12}\selectfont} 	
\usepackage{tabularx}
\usepackage[colorlinks]{hyperref}      
\hypersetup{linkcolor={blue},citecolor={blue},urlcolor={blue}}
\usepackage[normalem]{ulem}
\usepackage{amsfonts}       % blackboard math symbols
\usepackage{nicefrac}       % compact symbols for 1/2, etc.
\usepackage{microtype}  
\usepackage{fancyhdr}       % header
\usepackage{graphicx}       % graphics
\graphicspath{{media/}}   
\usepackage{multicol}
\usepackage{enumerate}
\usepackage[table, dvipsnames]{xcolor}
\usepackage{colortbl}
\renewcommand{\headrulewidth}{.4pt}

\usepackage{chemformula}
\usepackage{chemist}
\usepackage{chemfig}
\usepackage{graphicx}
\usepackage{afterpage}
\usepackage[utf8]{inputenc}
\usepackage{graphicx}
\usepackage{amssymb}
\usepackage{amsmath}
\usepackage{graphicx,wrapfig,lipsum}
\usepackage[capitalise]{cleveref} 
\Crefname{figure}{Fig.}{Figs.} % Capitalized version
\Crefname{equation}{Eq.}{Eqs.}
\usepackage{bm}
\usepackage{relsize}
\usepackage{float, multirow}
\usepackage{xcolor}
\usepackage{setspace}
\usepackage{mathptmx} % times new roman
\usepackage{newtxtext,newtxmath}
\usepackage{lscape}
\usepackage{soul} 
\usepackage{hanging}
\usepackage{xspace}
\usepackage[table]{xcolor}
\definecolor{R111G0B0}{RGB}{111,0,0}
\newcommand{\thead}[1]{\multicolumn{1}{c}{\textcolor{white}{\bfseries #1}}}

\usepackage{tikz}
\usetikzlibrary{positioning}
\usetikzlibrary{shapes.geometric, arrows}
\usepackage{calc}
\usetikzlibrary{arrows.meta}
\usepackage{varwidth}
\newcolumntype{?}{!{\vrule width 1.5pt}}
\usepackage[numbers, compress, square]{natbib}

\usepackage{listings} %for python code
\usepackage{tabulary}
\newcolumntype{L}[1]{>{\raggedright\let\newline\\\arraybackslash\hspace{0pt}}m{#1}}
\newcolumntype{C}[1]{>{\centering\let\newline\\\arraybackslash\hspace{0pt}}m{#1}}
\newcolumntype{R}[1]{>{\raggedleft\let\newline\\\arraybackslash\hspace{0pt}}m{#1}}
\usepackage[skip=2pt, font=small]{subcaption}
\makeatletter
\renewcommand{\fnum@figure}{Fig. \thefigure}	
\makeatother

\usepackage[dvipsnames]{xcolor}
\usepackage{sectsty}
\usepackage{booktabs}
\usepackage{array}      % for \newcolumntype
\usepackage{tabularx}   % for X/TabularX
\usepackage{siunitx}    % for S columns (number alignment)
\usepackage[table]{xcolor}

\definecolor{R111G0B0}{RGB}{111,0,0}

\newcolumntype{Y}{>{\raggedright\arraybackslash}X}

\usepackage{xcolor}
\definecolor{R111G0B0}{RGB}{111,0,0}

\usepackage{sectsty}
\sectionfont{\color{R111G0B0}}
\subsectionfont{\color{R111G0B0}}
\subsubsectionfont{\color{R111G0B0}}
\paragraphfont{\color{R111G0B0}}
\subparagraphfont{\color{R111G0B0}}

\usepackage[table]{xcolor}        % 'table' enables row coloring
\definecolor{R111G0B0}{RGB}{111,0,0}
\usepackage[toc,page]{appendix}
\usepackage{colortbl}             % ensure colortbl is loaded
\usepackage{array,tabularx}       % column types, TabularX
\usepackage{booktabs}             % optional; if you use \toprule/\midrule/\bottomrule

\usepackage{enumitem}

\usepackage{xcolor}
\definecolor{R111G0B0}{RGB}{111,0,0}

\usepackage[colorlinks]{hyperref}
\hypersetup{linkcolor=R111G0B0,citecolor=R111G0B0,urlcolor=R111G0B0}

\usepackage{hyperref}
\hypersetup{
  colorlinks = true,
  linkcolor  = R111G0B0,
  citecolor  = R111G0B0,
  urlcolor   = R111G0B0
}

\usepackage[parfill]{parskip}
\usepackage{fancyhdr}
\usepackage{lineno}
\usepackage{stackengine}

\newlist{romanitem}{enumerate}{1}
\setlist[romanitem,1]{label=\textbf{Stage \Roman*}:, align=left, leftmargin=*}

\usepackage{colortbl}
\begin{document}

\title{\textcolor{R111G0B0}{\textbf{Kinetic-Mamba}: Mamba-Assisted Predictions of Stiff Chemical Kinetics}}

\maketitle
\fancypagestyle{alim}{\fancyhf{}\renewcommand{\headrulewidth}{0pt}\fancyfoot[R]{\today}\fancyfoot[L]{Preprint}}
\thispagestyle{alim}
\vspace{-2cm}
\begin{center}
{\textcolor{R111G0B0}{\textbf{Additi Pandey}$^{a,*}$, \textbf{Liang Wei}$^{b}$, \textbf{Hessam Babaee}$^{c}$, \textbf{George Em Karniadakis$^{a}$}}}
\end{center}
\vspace{0.5cm}
\begin{center}
     \textsuperscript{a} Division of Applied Mathematics, Brown University, Providence, RI, USA \\
     \textsuperscript{b} Karagozian \& Case, Inc., Glendale, CA, USA \\
    \textsuperscript{c} Department of Mechanical Engineering and Materials Science, University of Pittsburgh, Pittsburgh, PA, USA
\end{center}
\vspace{0.5cm}
{\textcolor{R111G0B0}{\begin{abstract}
{\textcolor{black} 
{
Accurate chemical kinetics modeling is essential for combustion simulations, as it governs the evolution of complex reaction pathways and thermochemical states. In this work, we introduce Kinetic-Mamba, a Mamba-based neural operator framework that integrates the expressive power of neural operators with the efficient temporal modeling capabilities of Mamba architectures. The framework comprises three complementary models: (i) a standalone Mamba model that predicts the time evolution of thermochemical state variables from given initial conditions; (ii) a constrained Mamba model that enforces mass conservation while learning the state dynamics; and (iii) a regime-informed architecture employing two standalone Mamba models to capture dynamics across temperature-dependent regimes. We additionally develop a latent Kinetic-Mamba variant that evolves dynamics in a reduced latent space and reconstructs the full state on the physical manifold. The accuracy and robustness of Kinetic-Mamba was evaluated using both time-decomposition and recursive-prediction strategies. We further assess the extrapolation capabilities of the model on varied out-of-distribution datasets. Computational experiments on Syngas and GRI-Mech 3.0 reaction mechanisms demonstrate that our framework achieves high fidelity in predicting complex kinetic behavior using only the initial conditions of the state variables.
}}
\end{abstract}}}

\textcolor{R111G0B0}{}

\let\thefootnote\relax\footnotetext{\textcolor{R111G0B0}{* Corresponding author: additi\_pandey@brown.edu (Additi Pandey) }}

\section{Introduction\label{sec:introduction}} \addvspace{10pt}

% Combustion remains a socio-economically critical process, supplying more than 80\% of the global energy in various forms \citep{swaminathan_2023_machine,IEA_2023_stat}. 
Modeling chemically reacting flows requires solving detailed chemical kinetic mechanisms that may involve dozens to thousands of species and reactions. The governing system of ordinary differential equations (ODEs) is typically highly stiff, characterized by the coexistence of vastly disparate time scales, from fast radical chain-branching reactions occurring on microsecond or nanosecond scales to slow recombination or diffusion-limited processes evolving over milliseconds or longer.

\par The stiffness in the system of ODEs introduces severe numerical difficulties. Explicit time-integrators require the timestep to match the fastest chemical timescale, making them prohibitively expensive, even though the physics of interest may evolve on far slower scales. Implicit methods, which are generally A-stable, are therefore widely used to solve stiff chemical kinetics. However, they require repeated Jacobian evaluations and nonlinear iterations, leading to very high computational cost \cite{pope_1997_in_situ_adaptive, kee_2005_chemReactFlow}. Stiff ODE systems remain challenging even with advanced implicit techniques. Several methods have been developed to alleviate the algebraic burden while maintaining A-stability  \citep{Aro_1996_CHEMSODE, Aro_1996_Stiff, Aro_1999_High, Dabdub_1995_Extrapolation}, yet these approaches are often limited in their ability to extrapolate solutions over long physical times. Linearly implicit Rosenbrock methods require solving only a linear system, significantly reducing the overhead in solving non-linear equations from the implicit solvers \cite{Rang_2015_Rosenbrock}. Nevertheless, for reaction mechanisms involving hundreds and thousands of species, the resulting linear systems can still be extremely large, and solving them remains a major bottleneck.

Despite decades of advances in stiff ODE solvers, the efficient integration of detailed chemical kinetics remains a major challenge, particularly in turbulent reacting-flow simulations in computational fluid dynamics (CFD).  This strongly motivates the development of innovative acceleration strategies, including machine-learning-based surrogate models, to reduce the computational overhead associated with stiff chemical source-term evaluations while retaining physical fidelity. In this regard, several machine learning models have been proposed such as artificial neural networks (ANNs)-based models \cite{Sharma_2020_Deep} and  parallel ResNets \cite{Brown_2021_RESDON}. Physics-informed neural networks (PINNs) proposed by \cite{Raissi_2019_PINN} have also been used to accelerate stiff chemistry via quasi-steady state assumptions \cite{Ji_2021_Stiff} or through extreme theory of functional connections \cite{Frankel_2024_Hybrid}. Neural operators such as DeepONets \cite{Lu_2021_DeepONet}, FNOs \cite{Li_2020_FNO} and DeepOKANs \cite{Liu_2024_KAN, Shukla_2024_DeepOKAN} have also been widely used to accelerate stiff chemistry as the models can be trained offline, which takes the bulk of computational cost in consideration, and deployed online \cite{Goswami_2024_Stiff_Kinetics, Venturi_2022_SVDPF, Kumar_2024_Combustion, Weng_2025_ExFNO, Nath_2025_AMORE}. Most of these existing approaches, however, either rely on assigning a separate network to each variable or introducing a pre-transformation network within their model architecture, or depend on adaptive weighting schemes to handle variables and samples yielding higher errors.

% =========================================================
\par Since the equations that govern the dynamics of state variables are time-dependent ODEs, we can treat the trajectory of each state variable as a sequence in time. With the advent of transformer-based models, sequence modeling has been playing a major role in deep learning. However, the primary downside of employing transformers for tasks such as predicting the chemical kinetics of several state variables leads to substantial computational overhead due to the quadratic scaling with the sequence length. \citeauthor{Gu_2022_stSSM} \citep{Gu_2022_stSSM} introduced structured state space models (SSMs), which are foundational models of control theory. Through efficient reparameterization techniques introduced by \citeauthor{Gu_2022_stSSM}, structured state space sequence (S4) models have linear or near-linear scaling with sequence length to capture long dependencies, but showed promising results mostly on continuous data. Consequently, selective SSMs were introduced to model dynamical systems and learn long-range dependencies with linear sequence length. They employ a selection mechanism to parameterize the SSM parameters on the basis of the input. Combining the SSM architecture with the Multilayer Perceptron (MLP) transformer block produces Mamba \cite{Gu_2023_mamba} introduced by \citeauthor{Gu_2023_mamba} in 2023.
\citeauthor{Hu_2024_state} \citep{Hu_2024_state} has already touched on several aspects of modeling dynamical systems with Mamba, showcasing its efficiency and promising accuracy. While they demonstrate Mamba's ability to capture mildly stiff ODEs, the stiff chemical kinetics considered in our work, pose significant challenges that have not been addressed. 

\par Along with the promising applications of Mamba as language model, and its growing success in modeling dynamical systems, its usage in chemical and biological sciences is also increasing rapidly. \citeauthor{brazil2024mamba} \citep{brazil2024mamba} introduced a Mamba-based chemical foundation model trained on molecular data obtained from PubChem and demonstrated its efficient performance on varied tasks, including molecular property classification and prediction and prediction of chemical reaction yields. \citeauthor{Lombard_2024_molecular} \citep{Lombard_2024_molecular} explored the effectiveness of Mamba for molecular generation. \citeauthor{xu2024smilesmambachemicalmambafoundation} \citep{xu2024smilesmambachemicalmambafoundation} introduced SMILES-Mamba that leverages a combination of self-supervised pretraining on SMILES strings and fine tuning strategies to effectively capture the chemical structures as well as chemical relationships present within the molecular data. \citeauthor{xu2024proteinmambabiologicalmambamodels} \citep{xu2024proteinmambabiologicalmambamodels} proposed a Mamba-based architecture for protein function prediction by pretraining the model on protein amino acid sequence and fine-tuning it on specific protein functions. While these works demonstrate Mamba's capacity in dealing with chemical data, they do not address time-dependent predictions of stiff dynamical systems such as in chemical kinetics.

\par In this work, we extend the capabilities of Mamba-based models by applying them to accurately predict the dynamics of stiff chemical systems. We introduce Kinetic-Mamba framework that achieves comparable or better accuracy 
on similar mechanisms using single step training, without separate networks for each variable or adaptive weighting of the loss function.

\subsection{Contributions}
We present our framework Kinetic-Mamba (KM), which predicts the dynamics of the state variables involved in the reaction from the given conditions of state variables at a specific time point. Drawing on the approach of \citeauthor{Nath_2025_AMORE} \citep{Nath_2025_AMORE}, our model leverages time decomposition, extrapolation as well as recursive prediction to learn this mapping. Through this work, we want to demonstrate the effectiveness of selective SSMs like Mamba \cite{Gu_2023_mamba} in successfully modeling the stiff differential equations governing complex chemical kinetics. 
\par In this study, we present the following novelties: 
\begin{enumerate}
    \item We present Kinetic-Mamba (KM), a Mamba-based framework for predicting chemistry in stiff systems. Our Kinetic-Mamba framework consists of four different models:
    \begin{enumerate}
        \item A single Mamba model that predicts the dynamics of all state variables from the given initial conditions.
        \item Two Mamba models that work together to address different ignition regimes present due to different initial conditions for temperature.
        \item A latent Mamba model with asymmetric input-output design, where the model receives $d$-dimensional projections as input 
        and predicts directly in the full $(m+1)$-dimensional state space, where $d<(m+1)$.
        \item A mass-conserving Kinetic-Mamba (KM) model to automatically enforce the constraint that the sum of mass fractions of all the species equals to unity.
    \end{enumerate}
    \item Our Kinetic-Mamba model is able to predict the dynamics of state variables on variable length time windows irrespective of being trained on windows of fixed temporal length, enabling direct integration with adaptive CFD time-stepping schemes.
    \item We demonstrate the accuracy of our Kinetic-Mamba (KM) framework not only on the dataset generated using parameters sampled from a uniform distribution but also on the dynamics of state variables generated using pressure, temperature and equivalence ratios sampled from a random distribution. This allows us to perform a more realistic and holistic assessment of the prediction accuracy of our proposed framework.
\end{enumerate}

\section{Methodology\label{sec:method}} \addvspace{10pt}
Mamba is an extension of the structured state-space model (S4) \cite{Gu_2023_mamba}. These models utilize the following equations to map input sequence $x(t)$ into the output sequence $y(t)$ via hidden state $h(t)$:
\begin{subequations}
\label{Equations:Mamba_continuous}
\begin{align}
h^\prime (t) &= \mathbf{A}h(t) + \mathbf{B}x(t), \\
y(t) &= \mathbf{C}h(t).
\end{align}
\end{subequations}
For an input $x\in \mathbb{R}^{\text{input}}$, the matrices $\mathbf{A,B,C}$ are in $\mathbb{R}^{\text{hidden} \times \text{hidden}}, \mathbb{R}^{\text{hidden} \times \text{input}}, \mathbb{R}^{\text{output} \times \text{hidden}}$, respectively, where $h \in \mathbb{R}^{\text{hidden}}$ and $y \in \mathbb{R}^{\text{output}}$ \citep{Gu_2023_mamba, Hu_2025_DeepOMambaSM}. For problems involving time discretization using the continuous time sequence to sequence mapping is not ideal, hence we discretize the above equations using zero-order hold:
\begin{subequations} \label{Equations: Mamba_discrete}
    \begin{align}
        h_t = \mathbf{\bar{A}}h_{t-1}+\mathbf{\bar{B}}x_t,\\
        y_t = \mathbf{C}h_t,
    \end{align}
\end{subequations}
such that $\mathbf{\bar{A}}=f(\Delta, \mathbf{A})=\exp(\Delta \mathbf{A})$ and $\mathbf{\bar{B}}=f(\Delta,\mathbf{A},\mathbf{B}) = \exp(\Delta \mathbf{A})^{-1}(\exp(\Delta\mathbf{A})-\mathbf{I}) \cdot (\Delta \mathbf{B})$, where $\Delta, \mathbf{A}, \mathbf{B}$ are the continuous parameters. The continuous and the discrete parameters are linear time invariant, that is, they are fixed for all time steps. Mamba overcomes this by introducing a selection mechanism, by making $\mathbf{B}, \mathbf{C}$ and $\Delta$ as functions of the current input. Combined with hardware-aware algorithms (involving kernel fusion, parallel scan algorithm, and recomputation to reduce memory requirements), Mamba offers linear time efficiency, making it adept in sequence modeling. For stiff problems specifically, the discrete formulation closely resembles exponential integrators, which along with the dependency on current inputs, makes Mamba well-suited for stiff problems.

\par
\begin{figure}[htbp]
    \centering
    \includegraphics[width=\textwidth]{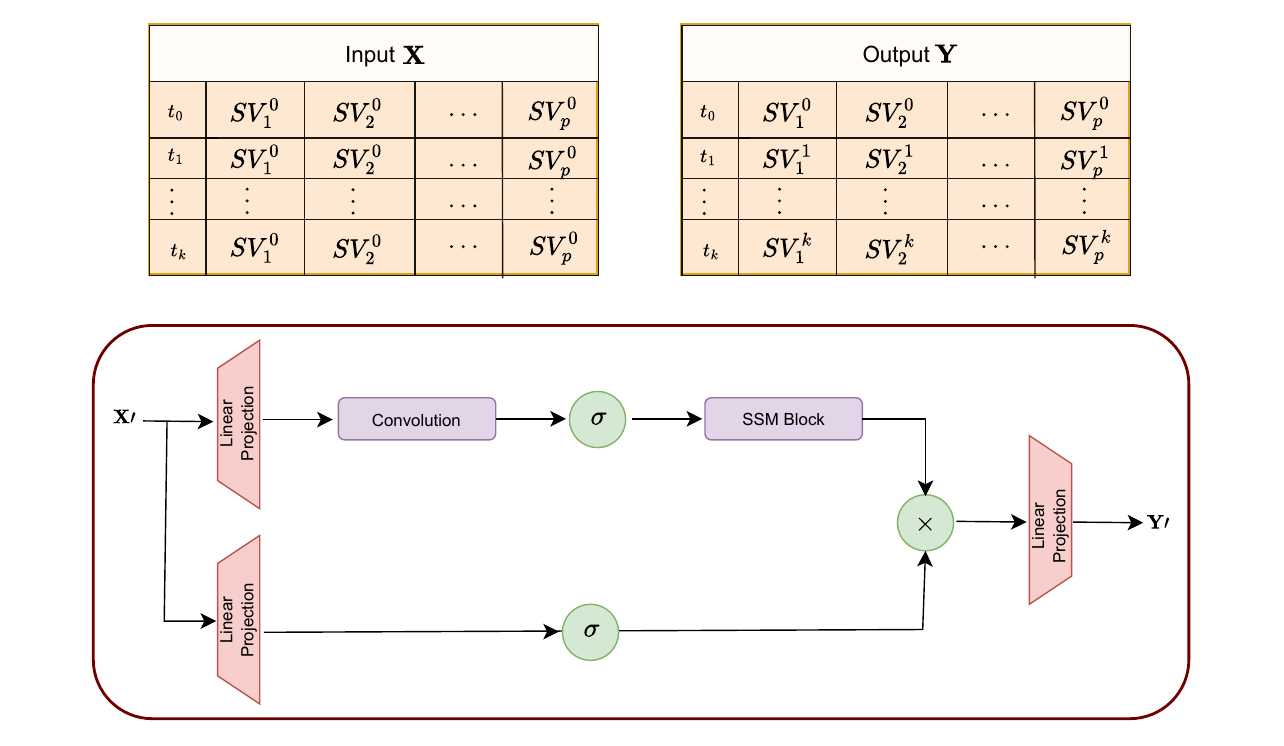}
    \caption{\textbf{Schematic representation of the Mamba block \cite{Gu_2023_mamba}}: The entire Mamba block consists of SSM related block and a skip connection block, combined via element-wise multiplication. The SSM related block consists of linear projection, convolution, non-linear activation function (SiLU) and SSM block, while the skip connection block consists of a linear transformation followed by non-linear activation. In the figure, the input $\mathbf{X}$ consists of initial values of state variables tiled to unknown time points, and $\mathbf{Y}$ has predicted values of state variables at unknown time points. These are denoted in the form of $SV_a^b$, where $a$ is the type of state variable (temperature, density, CO mass fraction, HO mass fraction, etc) and $b$ is the time. Moreover, $\mathbf{X}\prime, \mathbf{Y}\prime$ represent $\mathbf{X,Y}$ in $N$-dimensional hidden space after passing through the standard residual and norm layers.}
    \label{fig:mambablock}
\end{figure}

 In this study, we consider a deep sequence backbone of repeating Mamba-1 \citep{Gu_2023_mamba} blocks with linear input and output projection. We shall refer to it as a standalone Mamba model. To elucidate the model, consider that the input is given by $\mathbf{X} \in \mathbb{R}^{1 \times (k+1) \times p}$ which has initial conditions known at $t_0$ tiled across all time steps over which prediction is to be made. Although the initial condition is tiled across all time steps, temporal evolution is governed internally by the selective SSM blocks. We first project $p \rightarrow N$, where $N$ is the hidden dimension and process our input through $L$ blocks consisting of standard residual and normalization layers, followed by feeding the resulting state $\mathbf{X}\prime$ to the Mamba model. The output of the Mamba model $\mathbf{Y}\prime$ is processed through residual and normalization layers along with the application of MLP. After the final block, a final addition of residual and normalization is applied, after which the output is projected $N \rightarrow p$ dimensions yielding $\mathbf{Y}$. Inside a single Mamba block as shown in \Cref{fig:mambablock}, there is first a linear projection that projects the $N$-dimensional hidden space onto a higher dimensional latent space. The input is split along the feature dimension. A $1$-dimensional convolution is applied over the temporal direction followed by a non-linear activation function to the first part. The output is then fed into the selective SSM block where it is evolved in time via SSM equations using parallel scan. On the second part, a non-linear activation function is applied and it is then combined with the output of SSM block via multiplication. The resultant output is finally projected back onto $N$-dimensional space.

The objective of this study is to provide an accurate prediction of the dynamics of the different state variables involved in a chemical reaction, governed by a system of stiff ordinary differential equations, given the initial conditions of these state variables. For a chemical reaction involving $m$ species, the state vector consists of $m+2$ state variables, making $p=m+2$, accounting for temperature as well as pressure or density that influence the reaction. This information depends on whether the reactor assumes constant volume or constant pressure. 

\subsection{Normalization Strategy and Input Construction \label{sec: Normalization and Input Construction}}

Suppose that our dataset consists of $N$ training samples, $M$ testing samples, $n_t=10,000$ total time steps, and $m+2$ thermodynamic state variables. Moreover, suppose $\Delta t = 10^{-7}$ seconds. We denote the training and testing datasets with $X_{tr} \in \mathbb{R}^{N \times n_t \times (m+2)}$ and $X_{te}\in \mathbb{R}^{M \times n_t \times (m+2)}$, respectively. Since the dataset has been generated using Cantera \cite{Goodwin_2017_Cantera} solver, it may result in some extremely small negative values in the species mass fractions due to numerical error. We therefore, substitute any negative value with $0$. Thus, our dataset lies in $\mathbb{R}^{s \times n_t \times (m+2)}_{\geq 0}$, where $s \in \{N,M\}$ depending on training or testing dataset. 

\par Since in a CFD-integrated setting, the solver provides initial condition at regular intervals, we emulate this by assuming that within an interval of $100 \Delta t$ seconds, we know the initial conditions coming from the computational fluid dynamics (CFD) solver. However, since the integration of the framework with a CFD solver is beyond the scope of this study, we assume that the initial conditions are known at every $10$ $\mu$s or k= 101st time point. We then construct our model input by performing time decomposition. To do so, we convert our dataset of $N$(or $M$) samples, $n_t$ time steps, and $m+2$ state variables into dataset consisting of $N * 99$ (or $M * 99)$ samples, $101$ time steps and $m+2$ state variables. We achieve this by segmenting our time domain into $99$ segments of length $101$ where the 101st time point overlaps with the first time point of next segment. We then translate all segments within $[0,100]$ time step range. The time decomposition scheme can be thought of as a mapping from $\mathbb{R}_{\geq 0}^{s \times n_t \times (m+2)}\rightarrow \mathbb{R}_{\geq 0}^{(s * 99) \times 101 \times (m+2)}$ which transforms both $X_{tr}$ and $X_{te}$, where $s \in \{N,M\}$ depending on training or testing dataset, respectively. The idea is to be able to learn the dynamics within small temporal windows and capture any stiffness locally that occurs in the dynamics of the state variables. After time decomposition on our dataset, we proceed to normalize it.

\par Different state variables involved in a chemical reaction have different magnitudes that vary widely from each-other. Due to this, we need to scale these values within a common range, say $[-1,1]$, in order to enhance the convergence of our Kinetic-Mamba framework, without leaving smaller species unresolved. Consider $X \in \mathbb{R}_{\geq 0}^{(s * 99) \times 101 \times (m+2)}$, where $s$ are the number of samples given by $N$ or $M$, depending on whether it corresponds to the training dataset or testing dataset. We first perform a power transformation given by $\tilde{X} = X^{1/5}$ inspired by \citep{Kumar_2024_Combustion}, compressing the dynamic range, while also accounting for zeros in the dataset. Corresponding to $t_0$ or the first time point, we obtain $\tilde{X}^0 \in \mathbb{R}_{\geq 0}^{(s * 99) \times (m+2)}$, which is the number of time-decomposed samples for different state variables at the given initial time. We find the minimum and maximum values from the training dataset $\tilde{X}_{tr}$ and $\tilde{X}_{tr}^0$ to perform min-max normalization over our training and testing datasets as well as their corresponding initial condition inputs. Using the following mapping corresponding to $\tilde{X}_{tr}$ and $\tilde{X}_{tr}^0$, respectively, we map our dataset to a compact interval $[-1,1]$:
\begin{equation}
    \mathcal{N}(\tilde{X}) = 2 \left(\frac{\tilde{X}-({\tilde{X}_{tr}})_{min}}{({\tilde{X}_{tr})}_{max}-({\tilde{X}_{tr}})_{min}}\right)-1, \; \text{where `tr' corresponds to the training dataset} 
\end{equation}
\begin{equation}
    \mathcal{N}(\tilde{X}^0) = 2 \left(\frac{\tilde{X}^0-({\tilde{X}_{tr}^0})_{min}}{({\tilde{X}^0_{tr})}_{max}-({\tilde{X}^0_{tr}})_{min}}\right)-1, \; \text{where $0$ corresponds to the time step $t_0$} 
\end{equation}
The minimum and maximum values have been computed for each state variable across all time steps for all time decomposed samples of the training dataset. Moreover, for the initial conditions of the training dataset, the minimum and maximum values have been computed along the sample direction. Once the datasets have been normalized, we construct the input (which consists of the initial conditions or data corresponding to $t_0$) on a grid. To do this, we construct a grid of $k=101$ time points, corresponding to $s * 99$ samples, where $s \in \{N,M\}$ depending on whether it is training or testing data, for all state variables. Since we only know input values corresponding to $t_0$ for different samples, we tile these values to the remaining time points, where the dynamics are unknown. The reason for this is to make the model aware of the time points where dynamics is to be predicted, while avoiding introducing any bias, as $0$ is a valid normalized value in the normalized dataset.

\subsection{Kinetic-Mamba Models \label{KM_models}}
In this section, we focus on the discussion of the various models that make up the Kinetic-Mamba framework. In this study, we shall refer to our principal model as the standalone Mamba model that consists of a deep sequence backbone of repeating Mamba blocks, with linear input and output projection. A similar model has been employed before by \citeauthor{Hu_2024_state} in \cite{Hu_2024_state}. The input to the model consists of tensors of shape $(s * 99, 101, m+2)$, where $s\in\{N,M\}$ denoting the training or testing samples, respectively, constructed by taking the initial condition at $t_0$ and tiling it across the remaining $100$ time steps as discussed in \Cref{sec: Normalization and Input Construction}.

\par When the dataset has all profiles for each variable exhibiting similar nature of evolution, one model can efficiently learn the dynamics. However, sometimes the evolution of temperature with time shows distinct ignition regimes in its dynamics. This could be due to the system being weakly reactive with negligible heat release for a specific temperature range, while beyond the range, the chain-branching reactions may intensify rapidly, producing a rapid temperature rise and a new steady state. In this case, there is a critical ignition threshold at a certain $\tau$ K. For the dataset considered in this study, we find the value of $\tau$ by observing that the profiles that are below the threshold $\tau$K, satisfy the following equation: Given $\epsilon>0$:
\begin{equation}{\label{finding_temp_bifurcation}}
\tau = \max_{i \in S_\epsilon} T_i(0),\; \;\; \; \; \; \; \; \; \; 
\mathcal{S}_\epsilon = \left\{i\left|\right.\frac{T_i(n_t)-T_i(1)}{n_t-1}<\epsilon, i=0,1,\dots,N_s\right\}
\end{equation}
where $N_s$ is the total number of samples, including both the training and testing datasets. We then partition our dataset into two subsets such that initial temperature values corresponding to values below and equal to $\tau$ form non-igniting regime and the remaining values form igniting regime. For each of these subsets, we create the corresponding training and testing datasets and then employ an independent standalone Mamba model on each of them. We refer to this setup as ignition regime-informed KM model.

\par Inspired by \cite{Nath_2025_AMORE}, we also introduce a mass-conserving Mamba model within the Kinetic-Mamba framework. The mass fractions of species is defined as the mass of the species divided by the total mass of all the species involved in the reaction. Hence, the sum of mass fractions of all the species should be equal to $1$. Since our predictions should at any instant of time satisfy this constraint, we implement the mass-conserving model within our Kinetic-Mamba framework. The mass-conserving model consists of a standalone Mamba model which automatically enforces the conservation of mass constraint. The conservation of mass constraint, which in this case, corresponds to the sum of all mass fractions equal to unity, is enforced automatically, by noting that the conservation of mass criterion is a linear constraint on the mass fraction. Given $m$ species, we could encode the mass fraction vector using a $(m-1)$-dimensional vector. Inspired by \cite{Karniadakis_2005_Spectral_book}, in $m$ dimensions, we define the forward mapping by:

\begin{equation}
    z_1, z_2, \cdots z_{m-1} = g(y_1, y_2, \cdots y_{m})
    \label{forwardmap}
\end{equation}

We call this function $g$, where $g: [-1,1]^m \rightarrow[0,1]^{m-1}$, such that $g(\bm{y})\mapsto g(\bm{z})$. This function $g$ is invertible and therefore, is bijective. Moreover, we have:
\begin{subequations}
\begin{align}
    z_k &= \frac{y_k}{1-\sum_{j=1, j \neq k}^{m-1} y_j }, \quad 1 \leq k \leq m-2\\
    z_{m-1} &= y_{m-1}
\end{align}
\end{subequations}

To find the inverse map ($g^{-1}( \cdot ) $), such that
\begin{equation}
    y_1, y_2, \cdots y_{m} = g^{-1}(z_1, z_2, \cdots z_{m-1})
\label{inversemap}
\end{equation}
We start from the forward map
\begin{equation}
    z_k = \frac{y_k}{d_k}, \quad  k = 1, \dots, m-2, \quad \mbox{where} \quad  d_k = 1 - \sum_{j=1, j \neq k}^{m-1} y_j,
\end{equation}
Using a simple substitution on $y_k$, we obtain a system of linear equations given by:
\begin{equation}
    d_k + \sum_{j \neq k}^{d} z_j d_j = 1, \quad \text{for } k = 1, \dots, m-2.
\end{equation}
which for simplicity can be rewritten as:
\begin{equation}
\bm{A} \bm{d} = \bm{b},
\end{equation}
where:
 \( \bm{d} = [d_1, d_2, \dots, d_{m-2}]^T \) is the vector of unknowns, \( \bm{b} = [1, 1, \dots, 1]^T \) is a vector of ones, and \( \bm A \) is a \( (m-2) \times (m-2) \) coefficient matrix defined as:
\begin{equation}
    \bm A_{kj} =
    \begin{cases}
        1, & \text{if } k = j, \\
        z_j, & \text{if } k \neq j.
    \end{cases}
\end{equation}
Once the linear system is solved, we can compute the original 
$m$-dimensional mass fraction vector as:

\begin{subequations}
\begin{align}
    y_k &= z_k d_k, \quad 1 \leq k \leq m-2\\
    y_{m-1} &= z_{m-1}\\
    y_m &= 1 - \sum_{k=1}^{m-1} y_k.
\end{align}
\end{subequations}
A detailed account, including the derivation of the same, can be found in \cite{Nath_2025_AMORE}. For our purpose, this implies that instead of training the model for $m$ species, density or pressure and temperature, we just need to train it for $(m+1)$ variables as our total number of variables reduces by a count of $1$.

\subsubsection{Latent Kinetic-Mamba model \label{subsec:LatentMamba}}
In realistic scenarios of complex reactions, there are multiple state variables whose dynamics needs to be taken into account for a successful simulation of chemical kinetics. In such situations, it is computationally expensive to train models for all state variables due to an increased number of trainable parameters. We therefore propose a latent Kinetic-Mamba framework. In the latent KM model, we perform PCA on the centered training and testing dataset by doing eigen-decomposition on the training dataset's covariance matrix; we reduce the feature dimension from $m+1$ to $d$, such that $d < m+1$. To demonstrate the performance of the latent framework, we only consider $m+1$ state variables instead of $m+2$, that is, we consider temperature and $m$ species. Moreover, our latent Kinetic-Mamba model takes input from the $d$-dimensional manifold and accurately predicts the dynamics of the state variables in the $(m+1)$-dimensional manifold, learning both the dynamics as well reconstruction in a single model.

\par To further elucidate this, consider $\mathbf{X}_{tr}, \mathbf{X}_{te}$ to be our training and testing datasets consisting of $m+1$ state variables, $M * 99, N * 99$ samples and $101$ time points. On normalization, we obtain the initial values as $X^0_{tr}$ and $X_{te}^0$ after dropping tilde for brevity. Post normalization, we have matrices of size $(N * 99) \times (m+1)$ and $(M * 99) \times (m+1)$, respectively. We define $\mathbf{X}^{c,0}_{tr} = \mathbf{X}^0_{tr} - \overline {\mathbf{X}^0_{tr}}$, where the mean is taken in the sample direction. We construct the covariance matrix given by
\[
\mathbf{\Sigma} = \frac{(\mathbf{X}_{tr}^{c,0})^T \mathbf{X}_{tr}^{c,0}}{(N * 99)-1}\]
We then perform eigen decomposition and obtain eigenvalues and the matrix of eigenvectors $\mathbf{V}$. The obtained matrix $\mathbf{V}$ which is an $(m+1)\times (m+1)$ matrix is orthonormal. We reduce its dimensions by truncating it into a $(m+1) \times d$ matrix $\mathbf{V_r}$. We obtain our principal component from $\mathbf{X}_{tr}^{c,0} \cdot \mathbf{V_r}$. Similarly, we center testing dataset by $\mathbf{X}^{c,0}_{te} = \mathbf{X}^0_{te} - \overline{\mathbf{X}^0_{tr}}$ and then obtain the principal component from $\mathbf{X}_{te}^{c,0} \cdot \mathbf{V_r}$
and denote them with $\mathbf{PC}^0_{tr},\mathbf{PC}^0_{te}$, respectively.
\par
\begin{figure}[htbp]
    \centering
    \includegraphics[width=\linewidth]{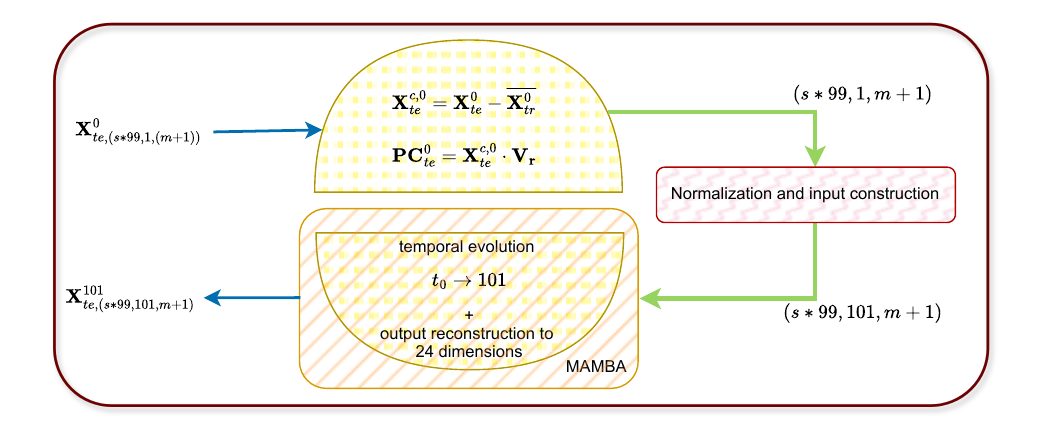}
    \caption{\textbf{Schematic representation of latent Kinetic-Mamba}: Corresponding to the GRI dataset, we input the initial conditions corresponding to 24 state variables (temperature, mass fractions of 23 active species) into our latent-KM model, and obtain their temporal evolution up to the 100th time step. Recall that we have already performed time decomposition on our dataset before we send it to the latent-KM model. Inside the latent-KM model, our state variables are transformed into latent variables, following the PCA procedure detailed above. They are then evolved by $n_t$ time steps and reconstructed by the Mamba block. This reconstruction happens due to the fact that our KM model is trained with output variables lying in the normalized time-decomposed 24-dimensional manifold.} 
    \label{fig:schem_KMPCA}
\end{figure}

 To train our Kinetic-Mamba model in latent space, we give as input the normalized latent variables $\mathbf{PC}_{tr}^0$, which lies in $d$-dimensional manifold after input construction (\Cref{sec: Normalization and Input Construction}). The normalization is done with respect to the minimum and maximum values of $\mathbf{PC}_{tr}^0$. We train the model using $\tilde{X}$ which lies in the $(m+1)$-dimensional manifold as output labels. This way, the model not only learns the dynamics in the latent space, but also reconstructs the output from $d$-dimensional manifold to the original $(m+1)$-dimensional manifold. \Cref{fig:schem_KMPCA} gives an overview of this methodology. 

\subsection{Training Methodology and Error Metrics \label{sec:training_error_method}}
We train our Kinetic-Mamba model using the time decomposition technique, as discussed in \Cref{sec: Normalization and Input Construction} with tiled initial conditions as input. The input to the Kinetic-Mamba is given on the time grid over which the prediction is expected. We train all our Mamba models within the Kinetic-Mamba framework using the mean squared error or MSE loss function with Adam \cite{Kingma_2014adam} optimizer and LambdaLR learning rate scheduler. The initial learning rate is set to $10^{-3}$ or $10^{-4}$, depending on the problem under consideration. We also assume the batch size to be $256$, unless stated otherwise. 

\par At inference, we obtain the predictions from the trained model given the initial conditions from the test dataset. We reconstruct the predictions across the time domain consisting of $99$ segments of $101$ time steps. The reconstruction is performed by concatenating the segments along the time dimension including common time points between any two segments. To evaluate the accuracy of the prediction, we then use the following metrics to compute the error between the prediction and the true values of the state variables.
\par Suppose that $y, \hat{y} \in \mathbb{R}_{\geq 0}^{(M \times (n_t-1) \times (m+2))}$ such that $y$ corresponds to the true values of the test dataset, while $\hat{y}$ corresponds to the predicted value, both reconstructed from the time decomposition scheme. There are three different error metrics we consider to compute the error in this paper: 
\begin{equation} \label{eq:L2time}
    \% \text{ relative $L_2$ error}_{(M,m+2)} = \left(\frac{||\hat{y}-y||_{2,t}}{||y||_{2,t}} \times 100 \right)
\end{equation}
\begin{equation} \label{eq:L2time_sample}
    \overline{\% \text{ relative $L_2$ error}}_{m+2} = \frac{1}{M}\sum_{j=1}^{M}\left(\frac{||\hat{y}-y||_{2,t}}{||y||_{2,t}} \times 100\right)
\end{equation}
\begin{equation} \label{eq:L2mean}
     \overline{\% \text{ relative $L_2$ error}} = \frac{1}{M \times (m+2)}\sum_{i=1}^{m+2}\sum_{j=1}^{M}\left(\frac{||\hat{y}-y||_{2,t}}{||y||_{2,t}} \times 100 \right)
\end{equation}
here $||.||_{2,t}$ denotes the $L_2$ norm along the temporal dimension consisting of $99 * 101$ time points. Moreover, the subscript$_{a,b}$ on LHS denotes the size of the resulting matrix. 

\par To calculate our error and plot the prediction within each segment for all segments in time decomposition setup, we plot each segment as a function of the global time within that segment and compute the relative $L_2$ error by flattening the time across all these segments to obtain $99 * 101$ time points for each state variable. In the Syngas problems, we omit plotting the dynamics corresponding to N$_2$ as it stays constant over time. We show our relative $L_2$ error values for all samples of each state variable over either a single or multiple independent runs in violin plots, with bulk of the violin indicating  that large number of samples have an error of around that value and tapering ends representing outlier samples with their corresponding error value on y-axis.

\subsection{Extrapolation \label{sec: extrapolation}}
It is challenging for models to generalize beyond the state variable range that they have been trained upon, specifically if there is stiffness within the solution profile. In this study, we shall also focus on assessing the accuracy of our framework in predicting the dynamics of the state variables generated with different equivalence ratio, temperature and pressure values outside the values they were trained upon. Since it is easier for models to generalize well on out-of-distribution samples generated using equivalence ratios and temperatures sampled from a uniform distribution as compared to a random distribution, we shall assess the performance of Kinetic-Mamba framework on out-of-distribution samples generated using equivalence ratios and temperatures sampled from a random distribution. Therefore, we shall generate different extrapolation data where each dataset corresponds to different equivalence ratio or temperature range, and pressure values or their combination. We shall then employ a pre-trained Kinetic-Mamba model to predict the dynamics of the state variables using initial conditions from the aforementioned extrapolation datasets. For a comprehensive assessment of the framework capabilities, our extrapolation dataset is such that it may show different ignition regimes. Therefore, utilizing \Cref{finding_temp_bifurcation} on the training dataset, we partition the dataset at the same critical temperature, perform normalization, construct the input as per \Cref{sec: Normalization and Input Construction}, and employ ignition regime-informed KM model consisting of the respective pre-trained models to predict the dynamics of the state variables. If the data lies completely below or above the critical ignition temperature, we only employ a single corresponding standalone KM model from the setup. 

\subsection{Recursive Predictions \label{sec:recursive_pred_method}}
Chemical kinetics often dominates the computational cost of reacting-flow CFD because stiff source terms must be evaluated at every cell and CFD timestep. To advance the thermochemical state over multiple CFD time steps, recursive prediction is employed, where the output of a pre-trained ML model is repeatedly fed back as input to the model. This approach integrates naturally with both fixed and adaptive CFD time-stepping. For fixed time steps, the CFD timestep can be viewed as a sliding window over the time domain, with each window encompassing a segment of fixed time points, e.g. 101 time points or 10 microseconds. For adaptive time steps, the recursive framework remains effective, as the model can be applied repeatedly over variable-length intervals, allowing the surrogate to track the kinetic trajectory accurately even when the CFD timestep changes dynamically. 
Using fixed timestep of 10 microseconds as an example, there are 99 segments over which we make predictions in the following fashion: 
\par Suppose the entire trajectory that needs to be predicted consists of $n_t$ time points from $t_0 = 0\Delta t$ seconds to $(n_t-1)\Delta t$ seconds. We start at $t_0$, construct our grid of $101$ time points by tiling the initial conditions as mentioned in \Cref{sec: Normalization and Input Construction} and then input it to the pre-trained model which predicts the values at those temporal grid points. To advance to the next window, we map the predicted value at the last time point of the last segment from the output normalization space to the branch input normalization space within the power-transformed space, via an inverse-forward transform, to account for different min-max transformations of branch input and model output space. We tile the result across the next $101$ time points and feed it into the pre-trained model. The process continues until we reach the last window containing the desired last time point.

\par An analogous process is applied when using latent KM model to predict recursively, with an additional step at each window boundary point. Herein, we take predictions of the last or rather the $101$-st time point of each window, which corresponds to $\mathbf{X}^{101}_{tr}$, apply an inverse-forward transform and obtain $\mathbf{PC}_{tr}^{101}$ by performing dimensionality reduction as outlined in \Cref{subsec:LatentMamba}. The resulting latent vector is tiled and fed to the pre-trained latent KM model for the next window. We repeat this process until we have predicted on the $(m+1)$-dimensional manifold up to all $9,901$ time points. However, when the time steps are not fixed, that is, when we have variable length temporal windows, we tile the initial conditions or the last point of the previous window up to the length of the desired next window. This adaptivity with Kinetic-Mamba model makes it robust for such predictions even though it was trained on segments or windows of fixed length $101$. Once we have obtained our predictions for each window, we concatenate them across the temporal dimension, removing repeated time points (corresponding to first point of prediction for all but first window) to obtain a complete trajectory on the given time domain.

\section{Computational Examples} 
We demonstrate the accuracy of our Kinetic-Mamba framework on two problems: the Syngas (synthetic gas) mixture and the GRI-Mech 3.0 \cite{GRI_mech} scheme. The dataset we obtained for Syngas is generated from two different distributions and hence, corresponding to this mechanism, we demonstrate the accuracy of our model on the two datasets for this mechanism by employing standalone Mamba, ignition regime-informed Mamba, and the mass-conserving Mamba model on it. To show the efficiency of latent Kinetic-Mamba model in accurately predicting the dynamics as well as reconstructing the dataset to the original $(m+1)$-dimensional manifold, we employ it to the GRI-Mech 3.0 dataset due to a higher number of species involved in the chemical reaction. To further test the capabilities of Kinetic-Mamba framework, we extrapolate on out-of-distribution samples for Syngas B dataset. We also perform recursive prediction on Syngas A as well as GRI-Mech 3.0 datasets.

\subsection{Time Decomposition\label{sec:time-decom-prob}}
\addvspace{10pt}

A Syngas problem was selected to develop, test, and demonstrate the workflow for building Kinetic-Mamba models for a stiff chemical kinetic mechanism. The case involves a skeletal Syngas mechanism for CO/H$_2$ combustion, consisting of 11 species and 21 reactions. This mechanism represents a key subset of larger hydrocarbon fuels. The dataset consists of the mass fractions of 11 species, temperature and pressure or density. The species under consideration are H$_2$, O$_2$, O, OH, H$_2$O, H, HO$_2$, CO, CO$_2$, HCO, N$_2$. We consider two datasets corresponding to this mechanism for analysis. In one configuration, pressure remains constant during the combustion process, while temperature and density undergo temporal evolution. Thus, for this case, the data was derived from a uniform distribution of temperature and equivalence ratios. In another configuration, the density remains constant while pressure and temperature undergo evolution with time. In this study, the data were derived from a random distribution of temperature and equivalence ratios within a constant volume reactor. In instances where temperature-pressure or temperature-density predictions are made along with the mass fraction of 11 species, the equation of state (EOS) can be employed to recover the density or pressure state variable. Therefore, the Syngas problem consists of 13 state variables. Henceforth, we shall refer to the aforementioned datasets as ``Syngas A" and ``Syngas B", depending on whether they are from the constant pressure or constant volume reactor, respectively.
The minimum and maximum values corresponding to the training and testing datasets are tabulated in \Cref{Appendix:DataRange}.

\subsubsection{Standalone Mamba Model \label{subsec: Problem1}}
\begin{figure}[htbp]
    \centering
    \includegraphics[width=\linewidth]{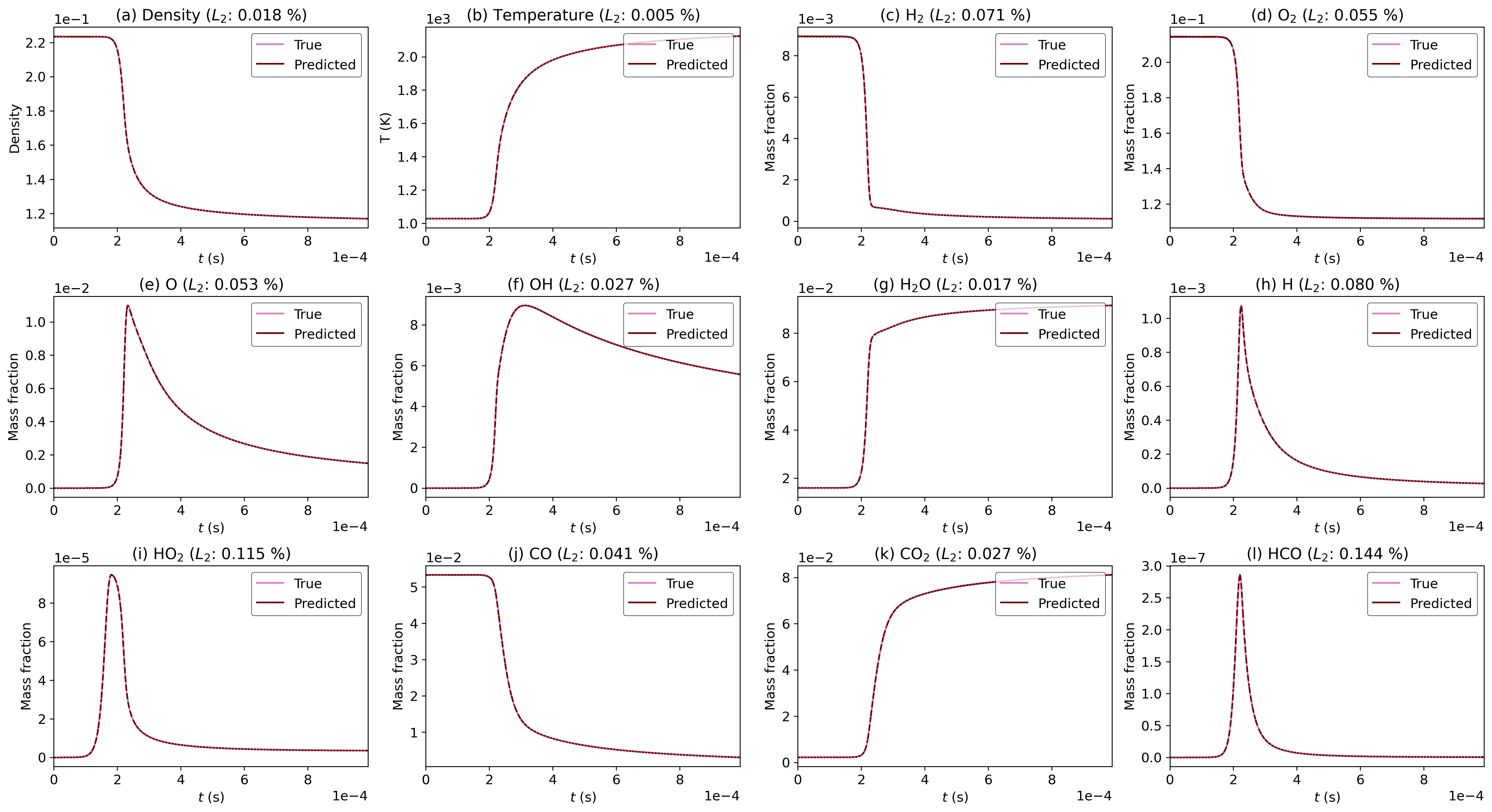}
    \caption{\textbf{Sample from Syngas A mechanism}: We can see good agreement between the predicted dynamics of the state variables and the ground truth values of the dynamics for an arbitrary test sample.}
    \label{fig:sample1_syngasA_vanilla}
\end{figure}

\begin{figure}[htbp]
    \centering
    \includegraphics[width=\linewidth]{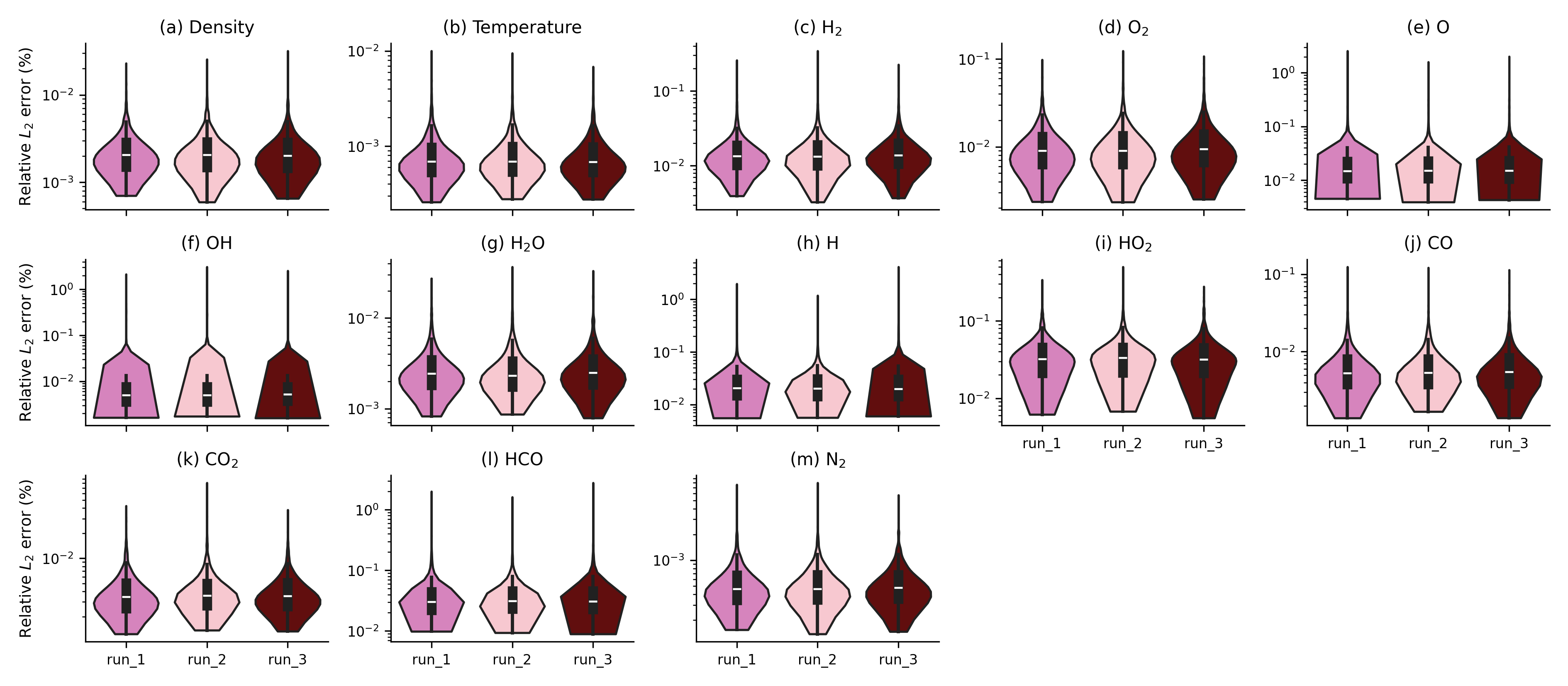}
    \caption{\textbf{Violin plot for the Syngas A mechanism:} This figure represents the percentage relative $L_2$ error obtained when the $L_2$ norm is taken with respect to the temporal dimension for both predicted and true test dataset. The plot in the log scale represents good accuracy across all samples for all 13 state variables.}
    \label{fig:vp_syngasA_vanilla}
\end{figure}

For a Syngas A model, we consider a syngas fuel composed of CO/H$_2$ with mole fractions CO = 0.5 and H$_2$ = 0.1, burned in air represented by an oxidizer mixture O$_2$ = 0.233 and N$_2$ = 0.767. The training data is generated using the Cantera library with an initial temperature in the range of [1000 K, 1500 K] and an equivalence ratio range of [0.7, 1.3]. The dataset consists of mass fractions of 11 species, temperature and density that vary across all time steps for all samples. The dataset also consists of pressure, which stays constant over time but has different values for different samples. For our model, we consider 13 thermochemical state variables comprising density, temperature, mass fractions of 11 species. Pressure at any instant for any sample can be recovered using the equation of state. Our training dataset consists of $6000$ samples across $10002$ time points for $13$ state variables, and testing dataset consists of $1500$ samples across $10002$ time points for the same state variables. We then scale this dataset onto $[-1,1]^{13}$ and perform time decomposition by segmenting the time domain into $99$ segments of temporal length $101$. The model is trained with $321,693$ parameters on the training dataset for 6001 iterations. For training the model, we use the Adam \cite{Kingma_2014adam} optimizer with LambdaLR learning rate scheduler and a learning rate of $10^{-3}$. We use MSE loss function and a batch size of $256$. \Cref{fig:sample1_syngasA_vanilla} shows the performance of the kinetic-mamba model on arbitrary test data sample. Across three runs, we reconstructed the output to the physical space over $99 * 101$ time points and computed the percentage relative $L_2$ error across all time steps according to \Cref{eq:L2time}. Corresponding to these runs, \Cref{fig:vp_syngasA_vanilla} shows these error values. The mean of this percentage relative $L_2$ error yielded the error corresponding to one run as given in \Cref{eq:L2mean}. Over three runs, we obtained an average error of $0.015\%$. The mean and standard deviation for all species is obtained by computing the mean of the percentage relative $L_2$ error across all samples (as per \Cref{eq:L2time_sample}) and then calculating its mean and standard deviation across three runs. We tabulate this information in \Cref{tab:vanilla_SyngasA}. 
\vspace{0.25cm}

\vspace{0.25em}
\begin{table*}[htbp]
\centering
\caption{\textbf{Syngas A Problem}:  Mean and standard deviation of percentage relative $L_2$ error values for Syngas A Mamba model across three independent runs.}
\label{tab:vanilla_SyngasA}
\renewcommand{\arraystretch}{1.25}
\setlength{\tabcolsep}{1.25pt}
  \begin{tabularx}{\textwidth}{@{}Y
      S[table-format=1.3e+2]
      S[table-format=1.3e+2]@{}}
    \rowcolor{R111G0B0}
\thead{State Variables} & \thead{$\mu$} & \thead{$\sigma$} \\ \toprule
\rowcolor{pink!45}\midrule
Density & 2.468e-03 & 2.040e-05 \\
Temperature & 8.703e-04 & 3.159e-06 \\
\rowcolor{pink!45}
H$_2$ & 1.685e-02 & 3.423e-04 \\
O$_2$ & 1.158e-02 & 2.754e-04 \\
\rowcolor{pink!45}
O & 2.558e-02 & 4.887e-04 \\
OH & 1.514e-02 & 2.417e-04 \\
\rowcolor{pink!45}
H$_2$O & 3.048e-03 & 6.947e-05 \\
H & 3.121e-02 & 1.357e-03 \\
\rowcolor{pink!45}
HO$_2$ & 3.720e-02 & 3.153e-04 \\
CO & 7.091e-03 & 6.028e-05 \\
\rowcolor{pink!45}
CO$_2$ & 4.423e-03 & 4.026e-05 \\
HCO & 4.388e-02 & 1.041e-03 \\
\rowcolor{pink!45}
N$_2$ & 5.826e-04 & 1.130e-06 \\ \bottomrule
\end{tabularx}
\end{table*}

\subsubsection{Mass-conserving Kinetic-Mamba Model \label{subsec: Problem2}}
\begin{figure}[htbp]
    \centering
    \includegraphics[width=\linewidth]{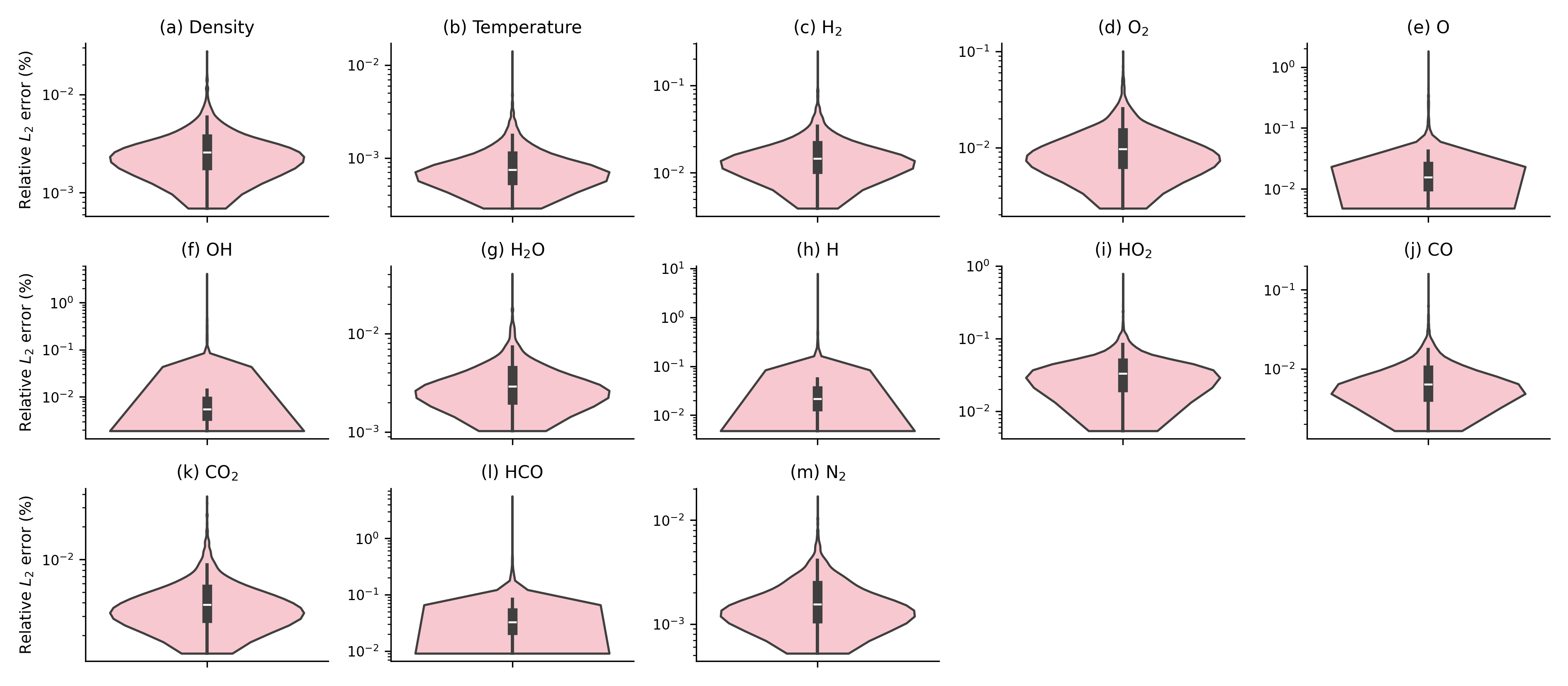}
    \caption{\textbf{Violin plot for the Syngas A mechanism using mass-conserving KM model:} This figure represents the percentage relative $L_2$ error obtained when the $L_2$ norm is taken with respect to the temporal dimension for both predicted and true test dataset. The plot represents good accuracy across all samples for all 13 state variables, as seen above in the log scale. The total relative $L_2$ error is 0.017\%.}
    \label{fig:vp_masscon}
\end{figure}

\begin{figure}[htbp]
    \centering
    \includegraphics[width=\linewidth]{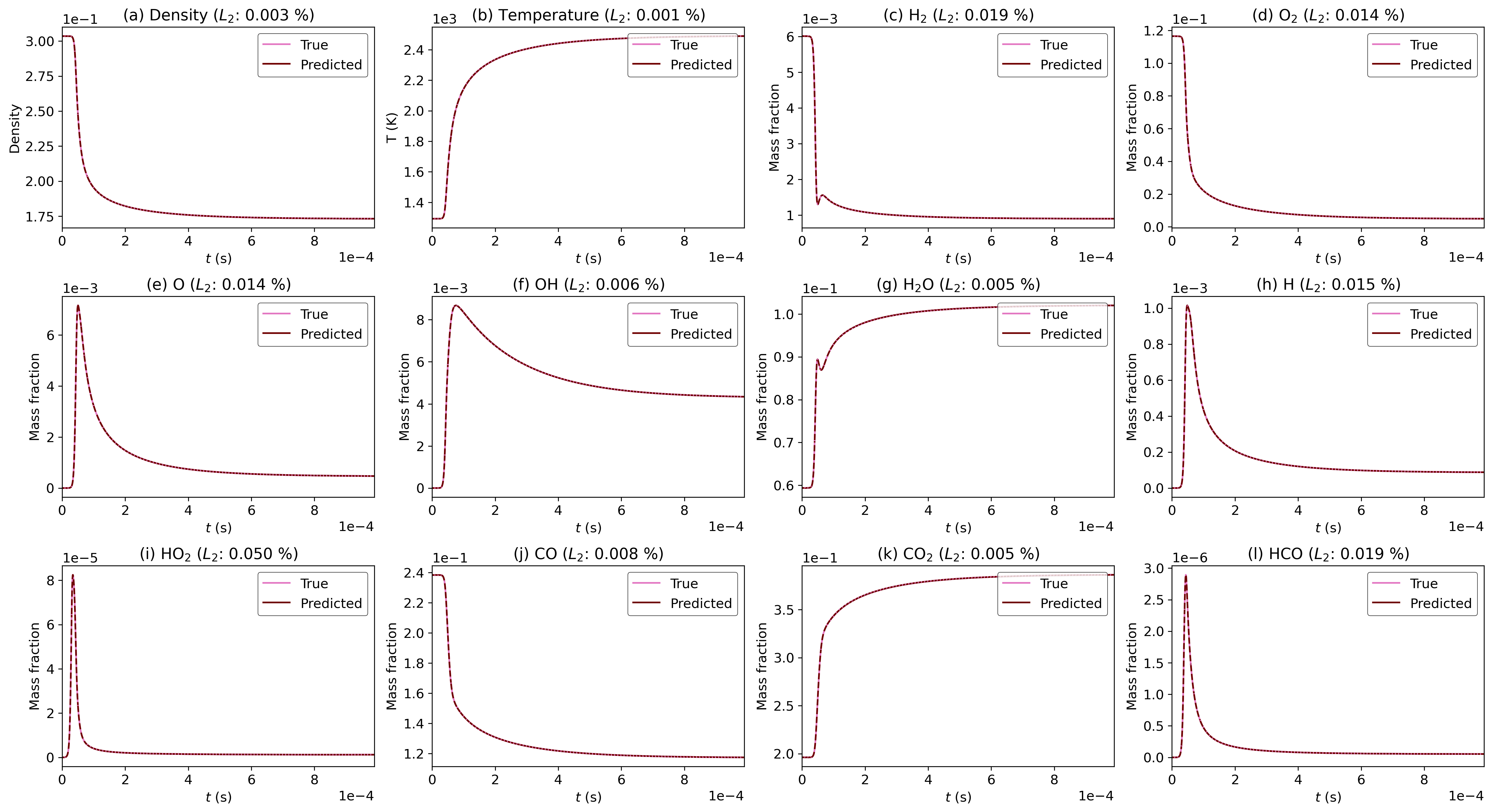}
    \caption{\textbf{Sample from Mass-conserving Kinetic-Mamba framework for Syngas A mechanism}: Plot of an arbitrary sample showing good agreement with ground truth values from the syngas A dataset when obtained using mass-conserving Kinetic-Mamba standalone model.}
    \label{fig:plot_masscon1}
\end{figure}

The sum of the mass fractions of all chemical species adds up to $1$ to ensure that the mass has been conserved over the course of the entire reaction. To do so, we enforce the conservation of mass constraint on 11 species involved in the reaction by projecting them onto a 10-dimensional manifold, bringing down the number of variables to $12$. This has been discussed in \Cref{KM_models}. We then feed the initial conditions of the 12 variables tiled over the time domain of $101$ time points, and train it with a standalone Mamba model using $321,532$ parameters. We use learning rate of $10^{-3}$ and LambdaLR learning rate scheduler with Adam \cite{Kingma_2014adam} optimizer for $6,001$ iterations. Upon reconstruction of the time domain, we plot the relative $L_2$ error across the time domain given by \Cref{eq:L2time} in \Cref{fig:vp_masscon}. We observed a mean relative $L_2$ error given by \Cref{eq:L2mean} across three runs as $0.017\%$. We tabulate the mean and standard deviation of the error values obtained using \Cref{eq:L2time_sample} for three runs in \Cref{tab:mss_conserving_syngasA}. We plot an arbitrary sample showing the predicted versus true values of the dynamics of state variables segment-wise in \Cref{fig:plot_masscon1}.

\vspace{0.25em}
\begin{table*}[htbp]
\centering
\caption{\textbf{Syngas A (Mass-Conserving) Problem}: This table shows the mean and standard deviation of the percentage relative $L_2$ error values over three runs for Mass-conserving Mamba model over the Syngas A dataset.}
\label{tab:mss_conserving_syngasA}
\renewcommand{\arraystretch}{1.25}
\setlength{\tabcolsep}{1.25pt}
  \begin{tabularx}{\textwidth}{@{}Y
      S[table-format=1.3e+2]
      S[table-format=1.3e+2]
      S[table-format=1.3e+2]
      S[table-format=1.3e+2]@{}}
    \rowcolor{R111G0B0}
\thead{State Variables} & \thead{$\mu$} & \thead{$\sigma$} \\ \toprule
\rowcolor{pink!45}\midrule
Density & 2.990e-03 & 8.715e-05 \\
Temperature & 9.335e-04 & 3.686e-06 \\
\rowcolor{pink!45}
H$_2$ & 1.820e-02 & 1.148e-04 \\
O$_2$ & 1.214e-02 & 1.629e-04 \\
\rowcolor{pink!45}
O & 2.624e-02 & 4.519e-04 \\
OH & 1.592e-02 & 1.006e-03 \\
\rowcolor{pink!45}
H$_2$O & 3.731e-03 & 9.375e-05 \\
H & 3.433e-02 & 1.186e-03 \\
\rowcolor{pink!45}
HO$_2$ & 3.905e-02 & 3.778e-04 \\
CO & 8.235e-03 & 1.118e-04 \\
\rowcolor{pink!45}
CO$_2$ & 4.876e-03 & 1.077e-04 \\
HCO & 4.715e-02 & 9.453e-04 \\
\rowcolor{pink!45}
N$_2$ & 1.945e-03 & 2.208e-05 \\ \bottomrule
\end{tabularx}
\end{table*}

\subsubsection{Ignition Regime-informed Kinetic-Mamba Model \label{subsec: Problem3}}
Our Syngas B fuel mixture contains 50\% CO, 10\% H$_2$, and 40\% N$_2$ by volume, while the oxidizer contains 25\% O$_2$ and 75\% N$_2$ by volume. This dataset is generated using a zero-dimensional constant-volume Cantera reactor. Initial equivalence ratios were randomly sampled from $\phi_0 \in [0.7, 1.4]$, and initial temperatures were randomly sampled from $T_0 \in [500, 2840]$ K. A total of $10,000$ samples were generated, each containing $10,000$ uniformly spaced time-points $(\Delta t = 10^{-7})$ s and the value at the first time point here is same as $\Delta t$. Additionally, it has temperature profiles that exhibit igniting vs non-igniting regimes during their course of evolution within the reaction.  
\par During the evolution of the mass fractions with time, temperature may often represent varied dynamics, such as certain profiles below a threshold remaining constant over time, while others increase and then reach a steady state. In short, above a certain threshold, called the critical ignition temperature, the temperature profiles may rise before plateauing, while below the threshold, the profiles remain flat. We refer to the two regimes as ignition and non-ignition regimes, respectively. We identify this threshold $\tau$ K using the formulation outlined in \Cref{finding_temp_bifurcation} with denominator equal to $9,999$. We split the dataset at $\tau$ into two parts, reflecting the two different ignition regimes. Corresponding to the initial temperature values below $\tau$K we have 1,400 samples in the training dataset and 350 samples in the testing dataset, while for the initial temperature values above $\tau$K, we have 6,600 training samples and 1,650 test data samples. In our case, the value of $\tau$K is roughly $889.09$K. 
\par
\begin{figure}[htbp]
    \centering
    \includegraphics[width=\linewidth]{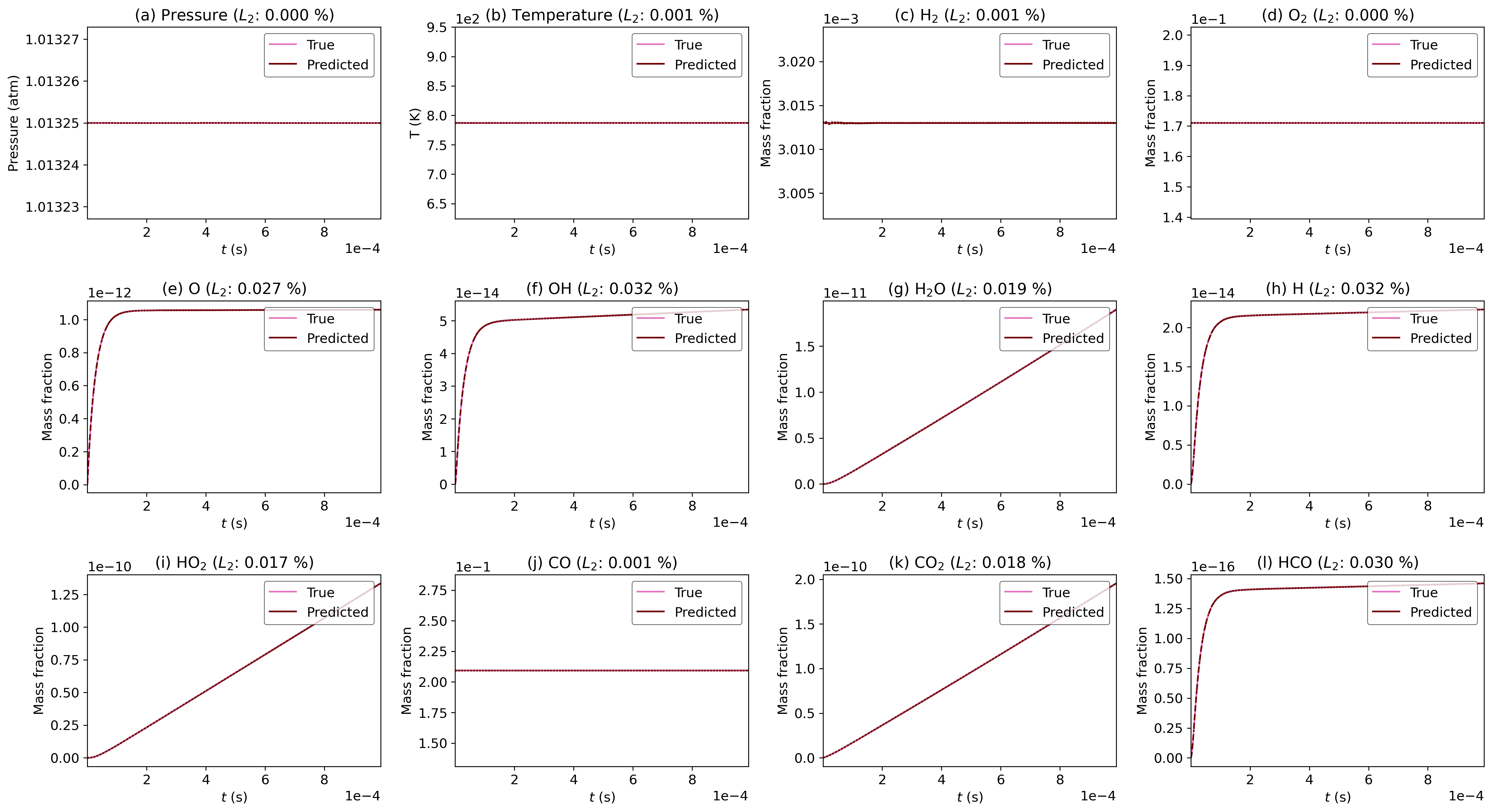}
    \caption{\textbf{Sample 1 from Syngas B mechanism}: We can see good agreement between the predicted dynamics of the state variables and the ground truth values of the dynamics for an arbitrary test sample lying in the non-ignition regime. We have expanded the y-axis in this figure to improve the visualization by removing small fluctuations due to constant values.}
    \label{fig:below_syngasB_sample}
\end{figure}

\begin{figure}[htbp]
    \centering
    \includegraphics[width=\linewidth]{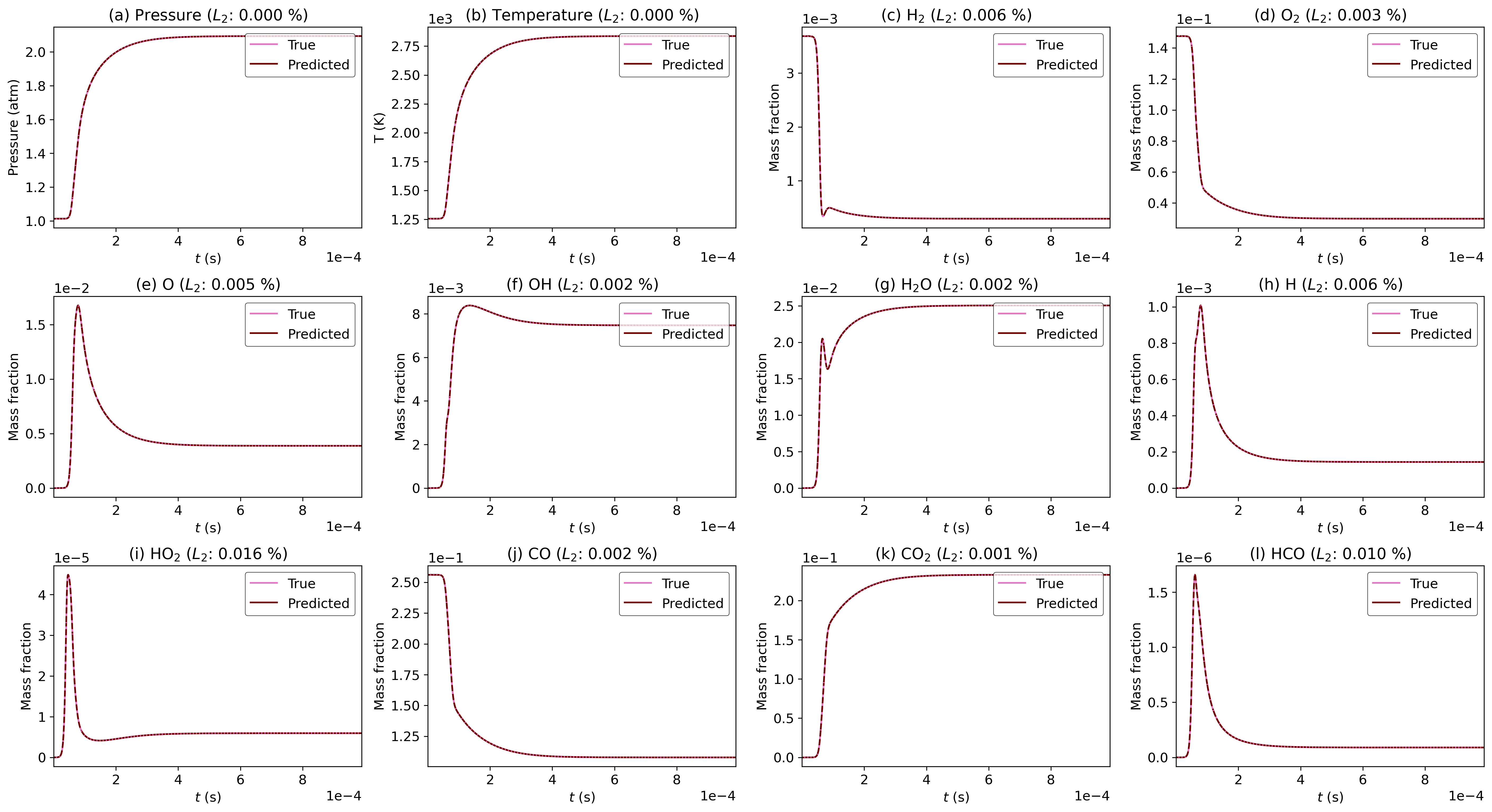}
    \caption{\textbf{Sample 2 from Syngas B mechanism}: We can see good agreement between the predicted dynamics of the state variables and the ground truth values of the dynamics for an arbitrary test sample lying within the ignition regime.}
    \label{fig:above_syngasB_sample}
\end{figure}

\begin{figure}[htbp]
    \centering
    \includegraphics[width=\linewidth]{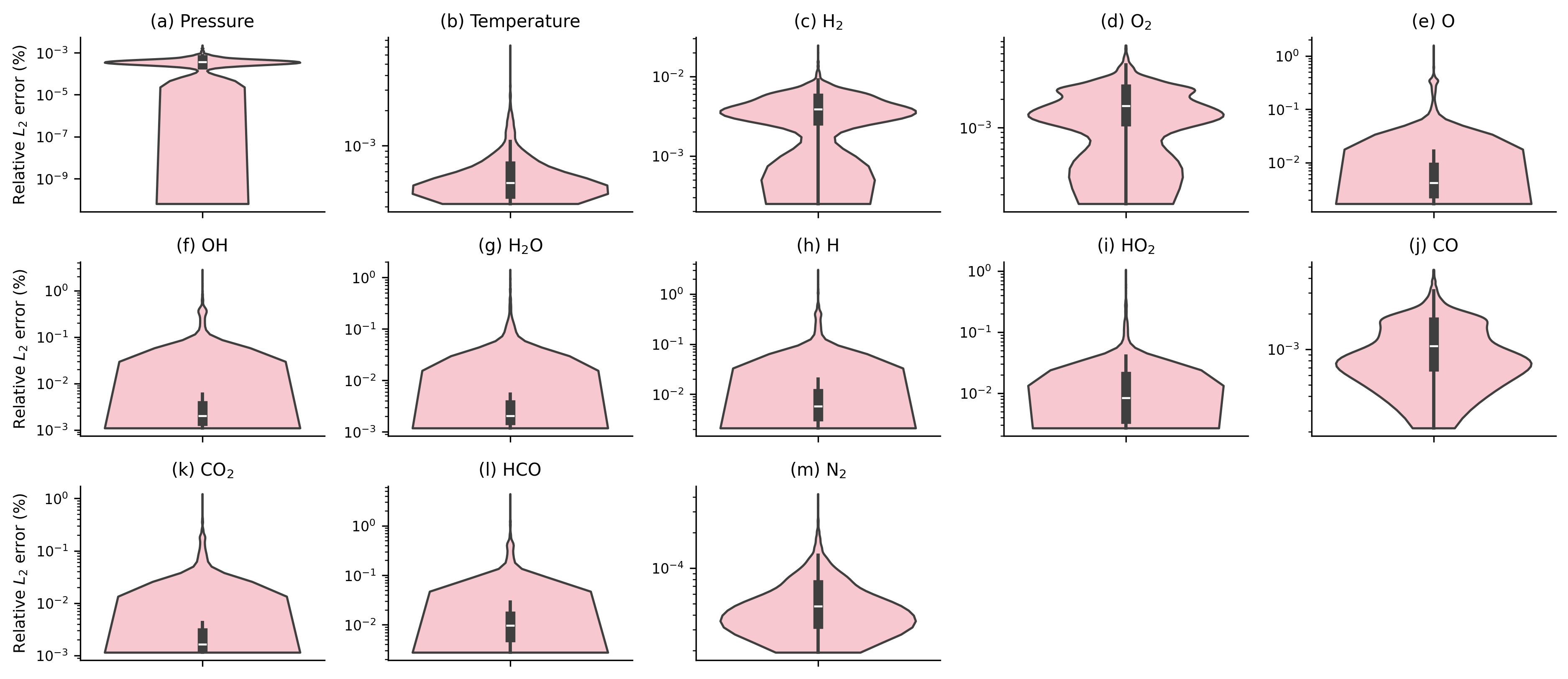}
    \caption{\textbf{Violin plot for the Syngas B mechanism:} This figure represents the percentage relative $L_2$ error obtained when the $L_2$ norm is taken with respect to the temporal dimension for both predicted and true test dataset. The plot is in log-scale on the y-axis. The plot refers to the Ignition Regime-informed Kinetic-Mamba standalone model which takes the ignition regime in the dataset into consideration.}
    \label{fig:syngasB_vp}
\end{figure}

 We train two standalone Mamba models with 231,757 and 384,943 parameters. Each model was trained using MSE loss criterion with learning rate $10^{-4}$, Adam \cite{Kingma_2014adam} optimizer and LambdaLR learning rate scheduler. The first model corresponding to constant temperature profiles was trained for 2,501 epochs with batch size of $256$, while the second model was trained for 5,001 epochs with a batch size of $256 \times 2$. The outputs from both the models were reconstructed to $99 * 101$ point time domain and then concatenated along the sample direction. We then computed the relative $L_2$ error across all time steps as given in \Cref{eq:L2time}. We show the true vs predicted values in the physical domain for temperature values below $\tau$K in \Cref{fig:below_syngasB_sample} and above $\tau$K in \Cref{fig:above_syngasB_sample}. The mean of the percentage relative $L_2$ error was taken, as per \Cref{eq:L2mean}, and average was computed across three runs. We obtained a total error of $0.023\%$. The mean and standard deviation of the error values for each state variable according to \Cref{eq:L2time_sample} is also tabulated in \Cref{tab:syngasb_errortable} for three independent runs. \Cref{fig:syngasB_vp} shows the relative $L_2$ error for all samples of each species when predictions were obtained from the Ignition Regime-informed Kinetic-Mamba model. We also observed that if this division in temperature profile was not performed to employ different standalone Mamba models, then the average error over three independent runs was $0.282\%$ when training was performed for 5,001 iterations with a batch size of $256$ and learning rate of $10^{-4}$ using the same optimizer and learning rate scheduler with 384,943 parameters.

\vspace{0.25cm}

\vspace{0.25em}
\begin{table*}[htbp]
\centering
\caption{\textbf{Syngas B Problem:} Mean and standard deviation of the percentage error values over three runs for Syngas B Mamba model.}
\label{tab:syngasb_errortable}
\renewcommand{\arraystretch}{1.25}
\setlength{\tabcolsep}{1.25pt}
  \begin{tabularx}{\textwidth}{@{}Y
      S[table-format=1.3e+2]
      S[table-format=1.3e+2]
      S[table-format=1.3e+2]
      S[table-format=1.3e+2]@{}}
    \rowcolor{R111G0B0}
\thead{State Variables} & \thead{$\mu$} & \thead{$\sigma$} \\ \toprule
\rowcolor{pink!45}\midrule
Pressure & 3.682e-04 & 4.816e-06 \\
Temperature & 5.952e-04 & 1.403e-05 \\
\rowcolor{pink!45}
H$_2$ & 3.946e-03 & 2.113e-05 \\
O$_2$ & 1.842e-03 & 1.872e-05 \\
\rowcolor{pink!45}
O & 4.709e-02 & 2.583e-03 \\
OH & 5.159e-02 & 3.900e-03 \\
\rowcolor{pink!45}
H$_2$O & 3.108e-02 & 4.360e-04 \\
H & 5.490e-02 & 4.066e-03 \\
\rowcolor{pink!45}
HO$_2$ & 2.977e-02 & 1.681e-03 \\
CO & 1.266e-03 & 4.753e-05 \\
\rowcolor{pink!45}
CO$_2$ & 2.490e-02 & 4.379e-04 \\
HCO & 5.861e-02 & 5.189e-03 \\
\rowcolor{pink!45}
N$_2$ & 5.814e-05 & 1.585e-07 \\
\bottomrule
\end{tabularx}
\end{table*}

\subsubsection{Latent Kinetic-Mamba Model \label{subsec:GRIProblem}} \addvspace{10pt}

\begin{figure}[htbp]
    \centering
    \includegraphics[width=\linewidth]{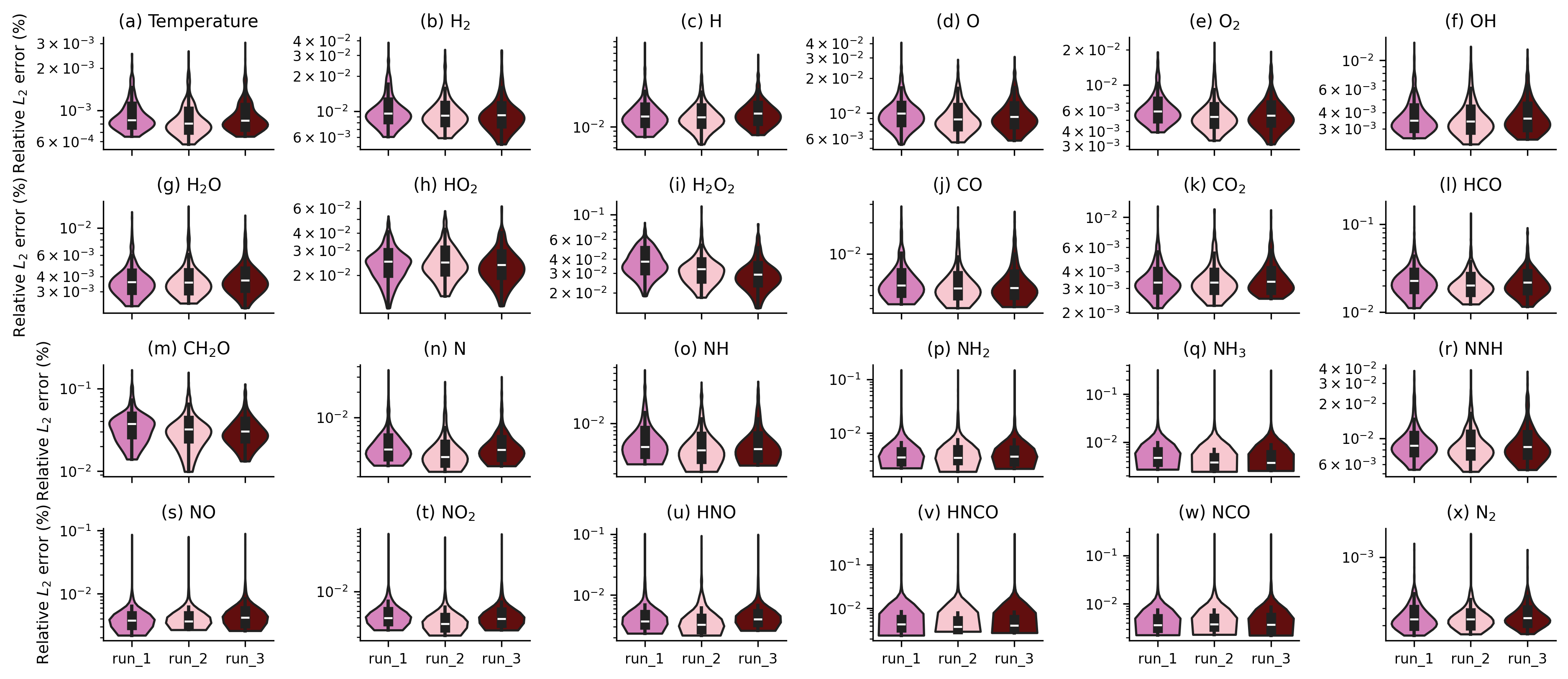}
    \caption{\textbf{Violin plot for the GRI mechanism:} This figure represents the percentage relative $L_2$ error obtained when the $L_2$ norm is taken with respect to the temporal dimension for both predicted and true test dataset. The plot represents good accuracy across all samples for all 24 state variables. (The plot has error values depicted in the log-scale on y-axis.)}
    \label{fig:vp_GRI}
\end{figure}

\begin{figure}[htbp]
    \centering
    \includegraphics[width=\linewidth]{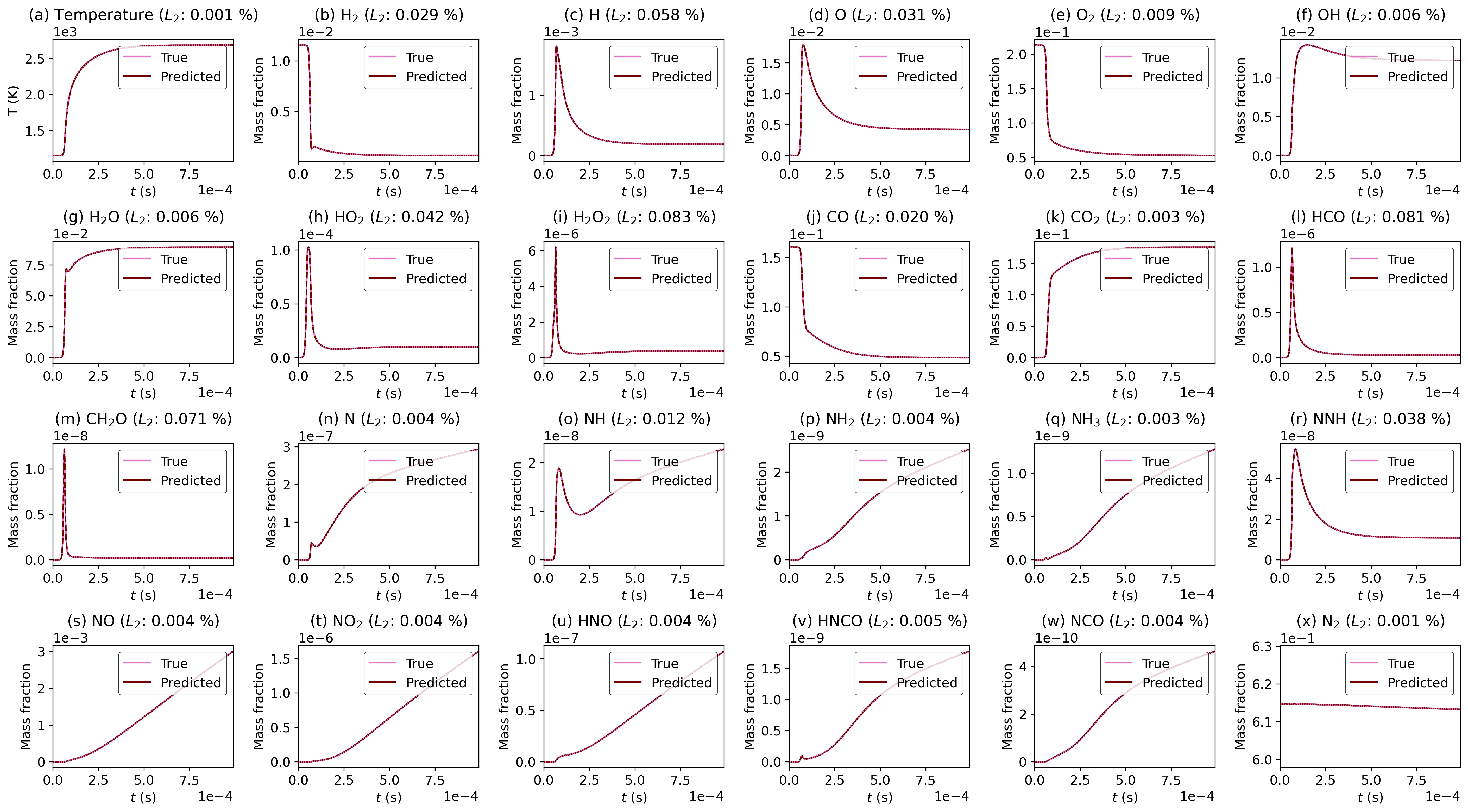}
    \caption{\textbf{Sample from GRI-Mech 3.0 mechanism}: We can see good agreement between the predicted dynamics of the state variables and the ground truth values of the dynamics for an arbitrary test sample.}
    \label{fig:sample_GRI}
\end{figure}

The GRI-Mech 3.0 \cite{GRI_mech} is an optimized mechanism for methane/air combustion. It consists of 53 species and 325 reactions. For the sake of our models, we only consider 23 species which have strictly positive mass fractions. Thus, we have 24 state variables including temperature and the 23 species' mass fractions. We partitioned the dataset into 2,080 samples of training dataset and 520 samples of testing dataset. 
In this problem as well, we divide our time series into 99 segments of length 101, that is, we perform time decomposition and then normalized the dataset as per \Cref{sec: Normalization and Input Construction}. We then converted our state variables into latent variables following \Cref{subsec:LatentMamba}. While we trained our model and predicted on latent manifold of latent dimension 12, we used output labels from the original 24-dimensional manifold, which were also normalized and time decomposed as per \Cref{sec: Normalization and Input Construction}. We trained the model with 232,408 parameters, batch size of 256 for 6,001 iterations using Adam \cite{Kingma_2014adam} optimizer and LambdaLR learning rate scheduler with learning rate $10^{-3}$. We computed the mean of the percentage relative $L_2$ error over the whole time domain of $101 \times 99$ time points as per \Cref{eq:L2mean} for three runs as $0.010\%$. We tabulate the mean and standard deviation of the errors computed using \Cref{eq:L2time_sample} for  three runs in \Cref{tab:errortable_GRI} and plot the percentage error values across all time steps for all samples of each state variable in \Cref{fig:vp_GRI}. We also plot the true values of the dynamics of state variables versus predicted values of the dynamics of the state variables obtained from our latent Kinetic-Mamba model in \Cref{fig:sample_GRI}.%and \Cref{fig:high_err_GRI}.

\vspace{0.25cm}

\begin{table*}[htbp]
\centering
\caption{\textbf{GRI-Mech 3.0 Problem}: Mean and standard deviation of the percentage error values over three runs for latent Kinetic-Mamba model.}
\label{tab:errortable_GRI}
\renewcommand{\arraystretch}{1.25}
\setlength{\tabcolsep}{1.25pt}
  \begin{tabularx}{\textwidth}{@{}Y
      S[table-format=1.3e+2]
      S[table-format=1.3e+2]@{}}
    \rowcolor{R111G0B0}
\thead{State Variables} & \thead{$\mu$} & \thead{$\sigma$} \\ 
\rowcolor{pink!45}\midrule
Temperature & 9.383e-04 & 2.768e-05 \\
H$_2$ & 1.062e-02 & 3.687e-04 \\
\rowcolor{pink!45}
H & 1.462e-02 & 4.583e-04 \\
O & 1.033e-02 & 3.490e-04 \\
\rowcolor{pink!45}
O$_2$ & 6.275e-03 & 2.855e-04 \\
OH & 3.935e-03 & 4.240e-05 \\
\rowcolor{pink!45}
H$_2$O & 3.969e-03 & 4.677e-05 \\
HO$_2$ & 2.563e-02 & 5.759e-04 \\
\rowcolor{pink!45}
H$_2$O$_2$ & 3.529e-02 & 3.523e-03 \\
CO & 5.911e-03 & 1.720e-04 \\
\rowcolor{pink!45}
CO$_2$ & 3.667e-03 & 3.392e-05 \\
HCO & 2.403e-02 & 1.095e-03 \\
\rowcolor{pink!45}
CH$_2$O & 3.707e-02 & 2.549e-03 \\
N & 4.926e-03 & 2.343e-04 \\
\rowcolor{pink!45}
NH & 6.796e-03 & 5.717e-04 \\
NH$_2$ & 5.207e-03 & 9.980e-05 \\
\rowcolor{pink!45}
NH$_3$ & 6.310e-03 & 4.782e-04 \\
NNH & 9.781e-03 & 1.869e-04 \\
\rowcolor{pink!45}
NO & 4.566e-03 & 2.943e-04 \\
NO$_2$ & 4.396e-03 & 3.848e-04 \\
\rowcolor{pink!45}
HNO & 4.859e-03 & 2.041e-04 \\
HNCO & 6.851e-03 & 1.475e-04 \\
\rowcolor{pink!45}
NCO & 5.590e-03 & 1.915e-04 \\
N$_2$ & 2.586e-04 & 5.212e-06 \\ \bottomrule
\end{tabularx}
\end{table*}

\subsection{Extrapolation}
\begin{figure}[htbp]
    \centering
    \includegraphics[width=\linewidth]{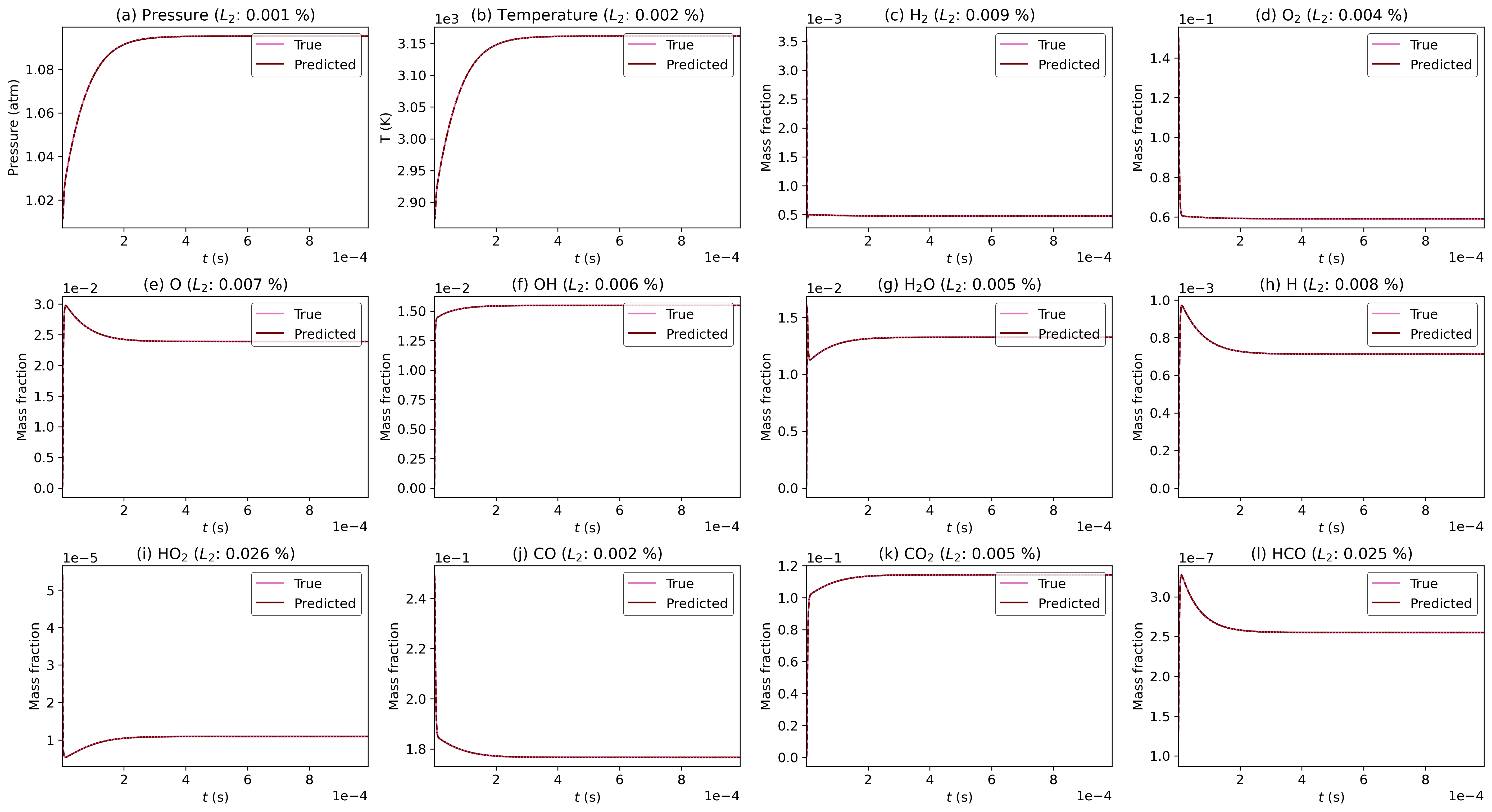}
    \caption{\textbf{Extrapolation Sample 1 for Syngas B}: Plot of predicted vs true data of an arbitrary sample from extrapolation dataset \Cref{extrap:case1}.}
    \label{fig:extra_plot_1}
\end{figure}

\begin{figure}[htbp]
    \centering
    \includegraphics[width=\linewidth]{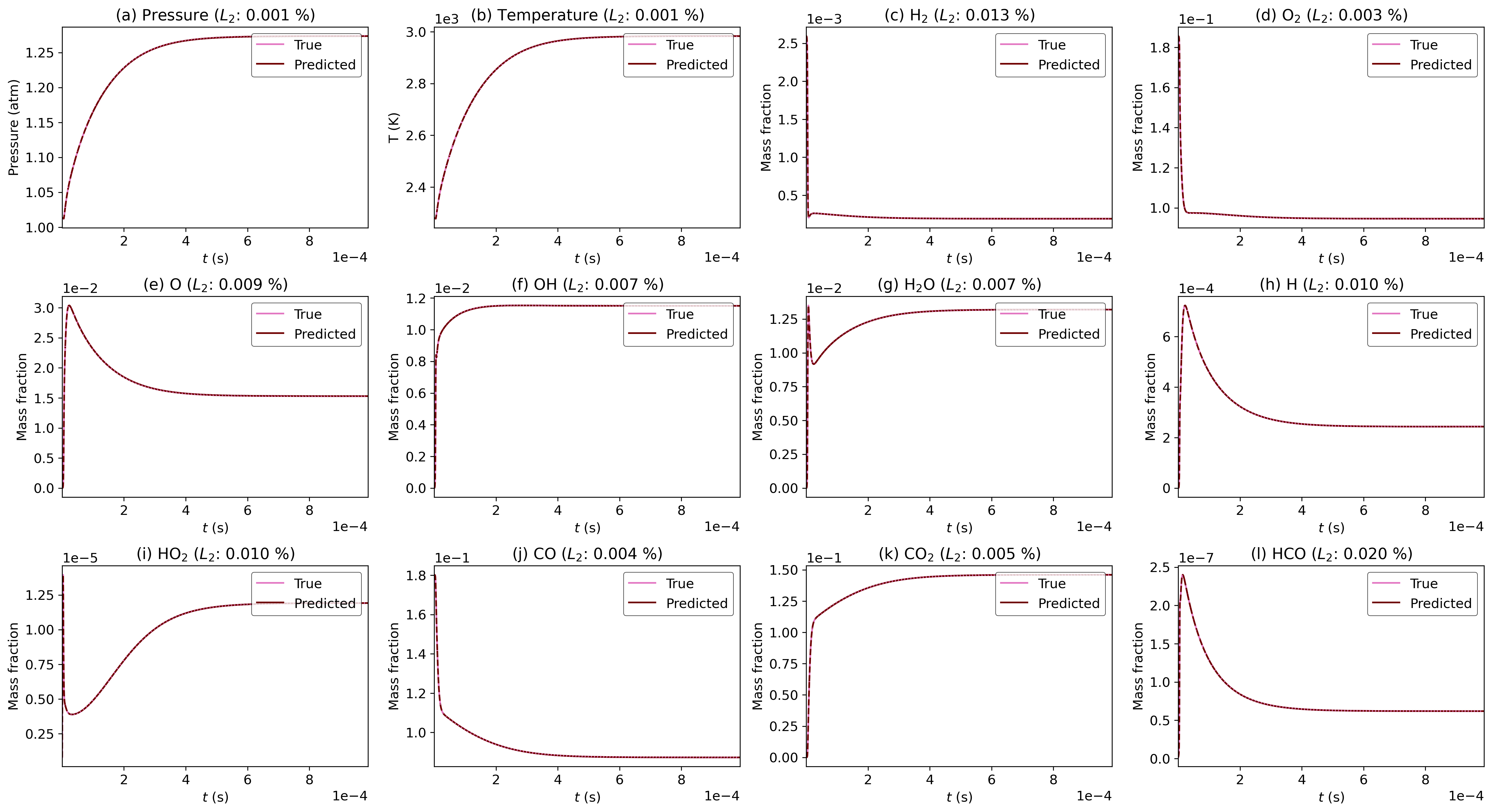}
    \caption{\textbf{Extrapolation Sample 2 for Syngas B}: Plot of predicted vs true data of an arbitrary sample from extrapolation dataset \Cref{extrap:case2}.}
    \label{fig:extra_plot_2}
\end{figure}

\begin{figure}[htbp]
        \centering
    \includegraphics[width=\linewidth]{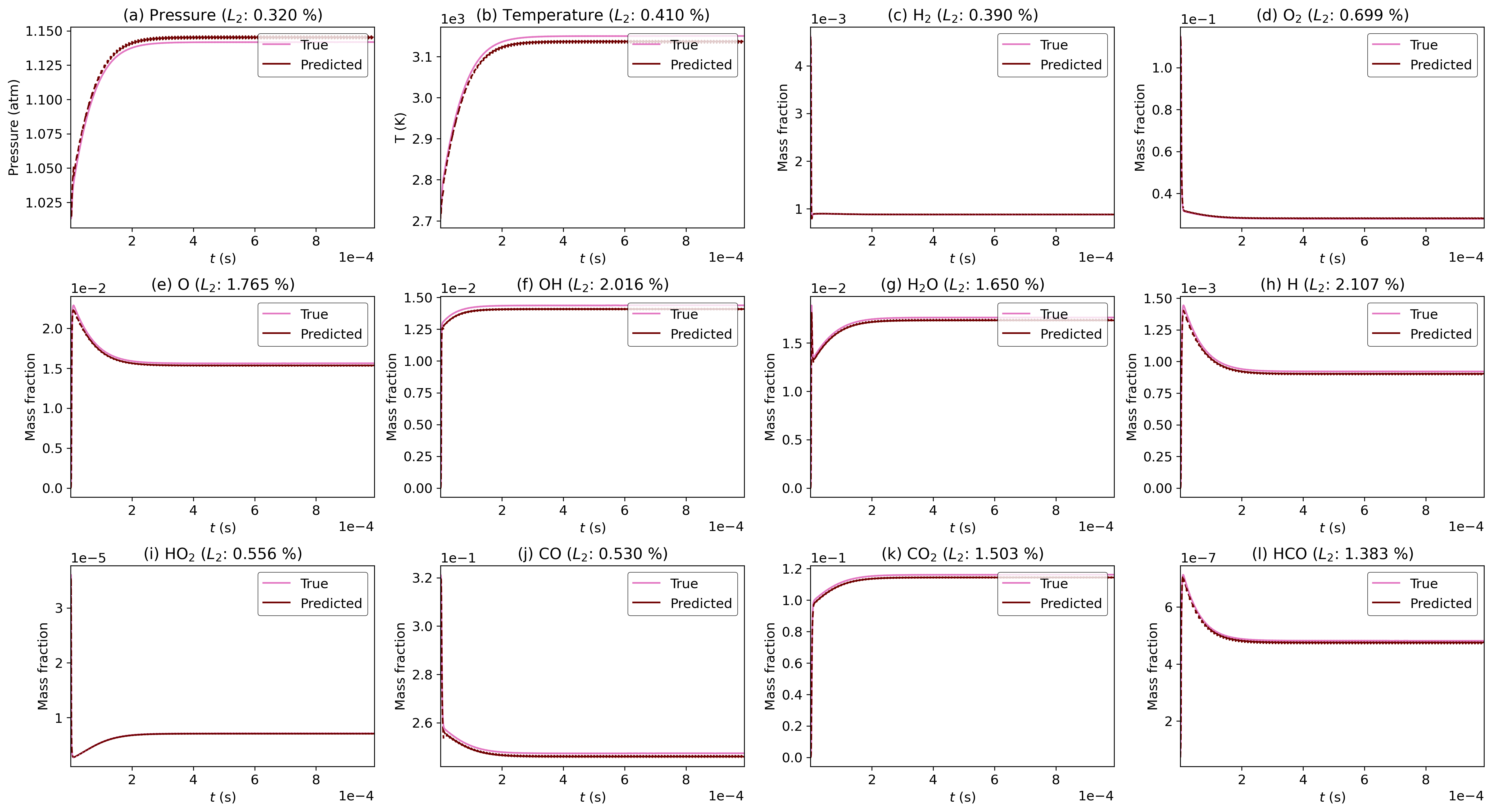}
    \caption{\textbf{Extrapolation Sample 3 for Syngas B}: Plot of predicted vs true data of an arbitrary sample from extrapolation dataset \Cref{extrap:case3}.}
    \label{fig:extra_plot_3}
\end{figure}

\begin{figure}[htbp]
    \centering
    \includegraphics[width=\linewidth]{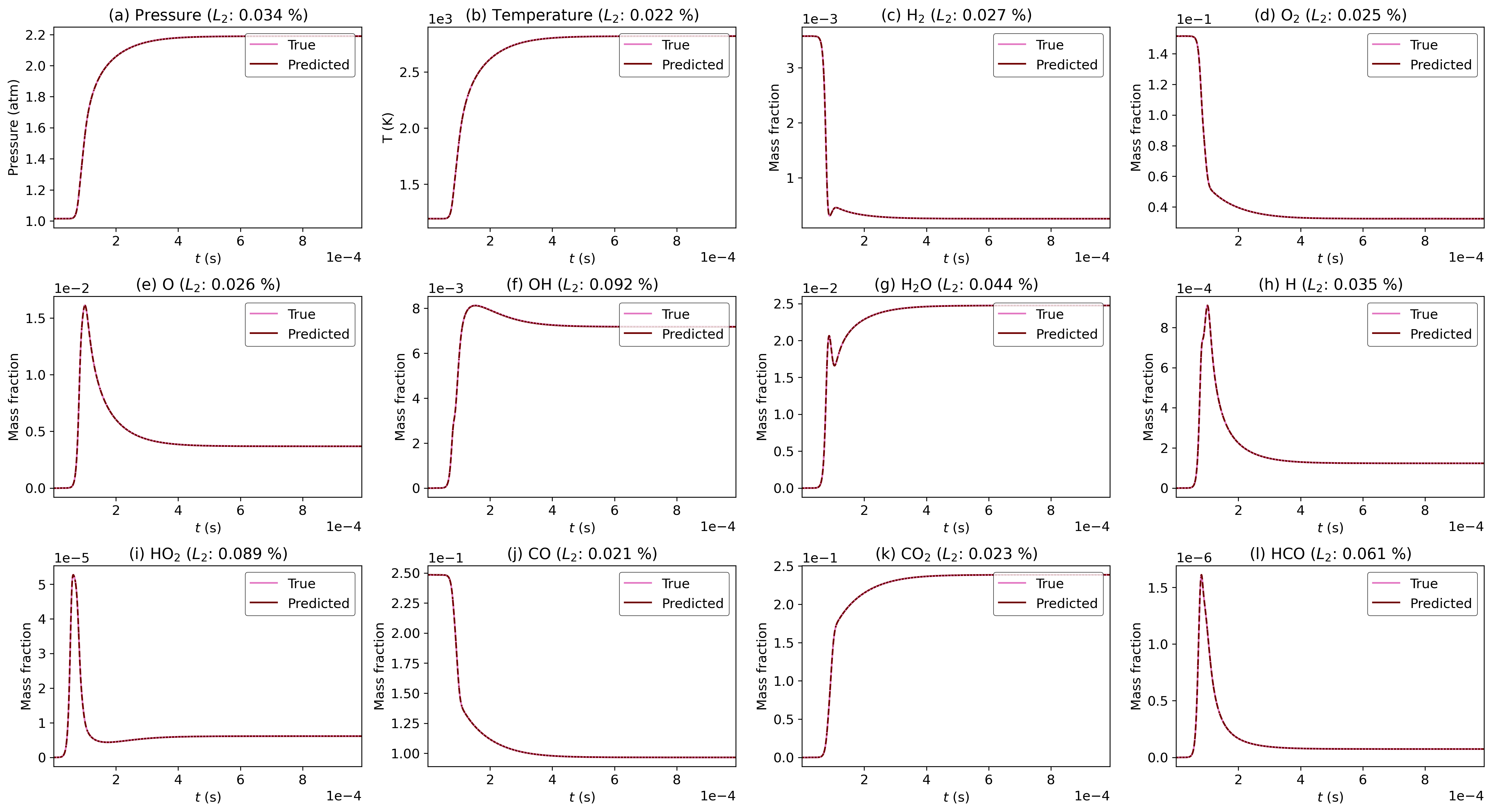}
    \caption{\textbf{Extrapolation Sample 4 for Syngas B}: Plot of predicted vs true data of an arbitrary sample from extrapolation dataset \Cref{extrap:case4}.}
    \label{fig:extra_plot_4}
\end{figure}

\begin{figure}[htbp]
    \centering
    \includegraphics[width=\linewidth]{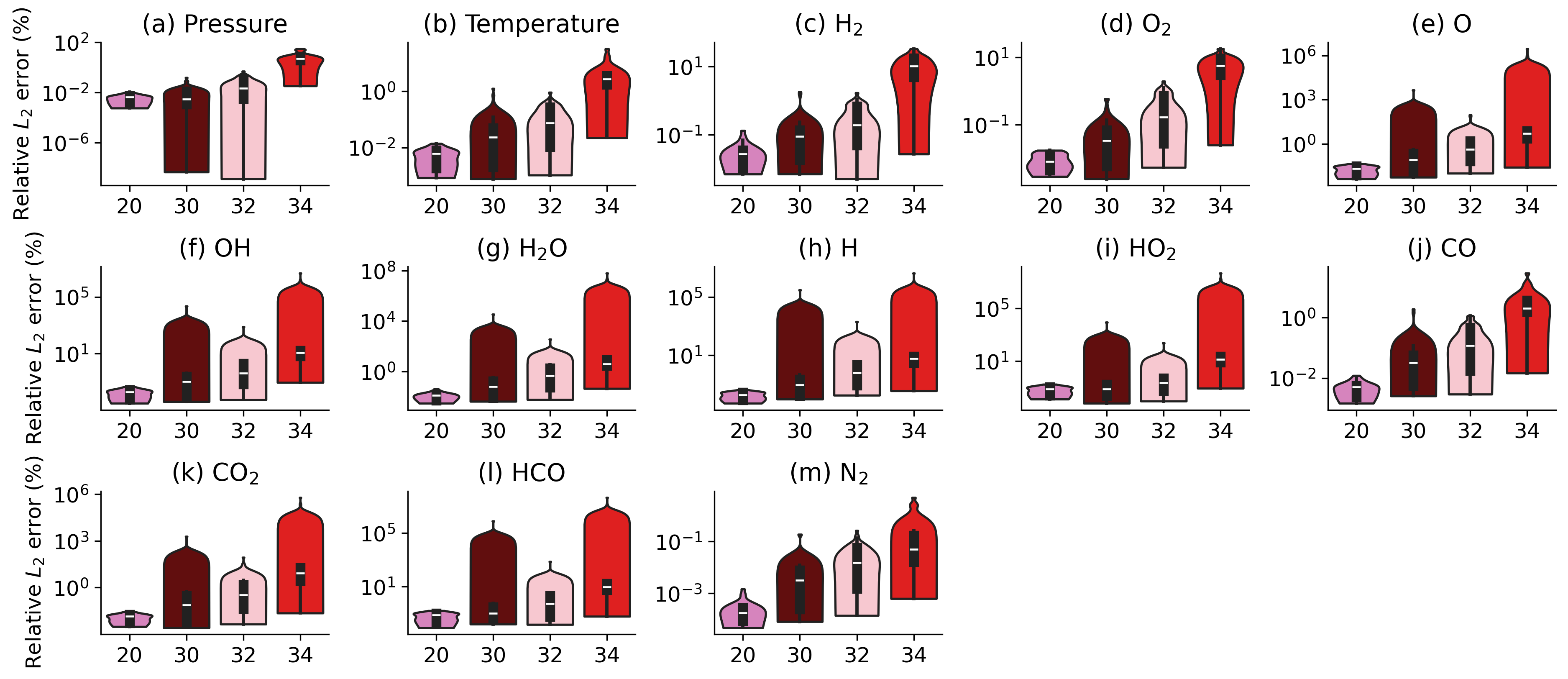}
    \caption{\textbf{Extrapolation for Syngas B KM model}: Violin plot obtained when pre-trained ignition regime-informed KM model is employed on extrapolation datasets \Cref{extrap:case1}, \Cref{extrap:case2}, \Cref{extrap:case3}, \Cref{extrap:case4} containing 20,30,32, and 34 samples respectively. Here, the error values are plotted in log-scale.}
    \label{fig:extrapolation}
\end{figure}

We discussed our method for performing extrapolation in \Cref{sec: extrapolation}. In this section, we shall further outline the method as well as provide results of performing extrapolation using ignition regime-informed KM model. We use our pre-trained model as defined in \Cref{KM_models} and then input the extrapolation datasets one by one, assuming same critical temperature as determined during the training process to assess the extrapolation accuracy. For this study, we generated four extrapolation datasets corresponding to the following initial temperature $T_0$, equivalence ratio $\phi$ and pressure $P$ values:
\begin{enumerate}
\item \label{extrap:case1} 20 samples with $T_0$ within $[2840, 3040]K, \phi \in [0.7, 1.4]$ and pressure of 1 atm.
\item \label{extrap:case2} 30 samples with $T_0 \in [500, 2840]K, \phi \in [0.5, 0.7]$ and pressure of 1 atm.
\item \label{extrap:case3} 32 samples with $T_0 \in [500,2840]K, \phi \in [1.4,2.0]$ and 1 atm pressure.
\item \label{extrap:case4} 34 samples with $T_0 \in [500, 2840]K, \phi \in [0.7,1.4]$ and pressure $\in [1, 1.5]$ atm.

\end{enumerate}
We plot arbitrary samples from these datasets in \Cref{fig:extra_plot_1}, \Cref{fig:extra_plot_2}, \Cref{fig:extra_plot_3}, \Cref{fig:extra_plot_4}, respectively. The violin plot \Cref{fig:extrapolation} shows the accuracy obtained when we extrapolated using pre-trained Kinetic-Mamba models discussed in \Cref{subsec: Problem2} for all samples of all the extrapolation datasets. We note that the model performs well for out-of-distribution temperature values but has some deterioration as the range equivalence ratio is changed. The performance worsens further with increasing pressure value but remains reasonable for most samples. Many species have very small values and sometimes the error value though small is overshadowed by an even smaller denominator $d$ (true value) during computation of relative $L_2$ error as per \Cref{eq:L2time} causing dramatic inflation in error values. We therefore, clip the denominator by $\max(\epsilon, d)$ and revise the error metric by clipping the denominator to get reasonable values as detailed in \Cref{appendix:additional_recursion_syngasA}.

\vspace{0.25cm}

\vspace{0.25cm}

\vspace{0.25cm}

\vspace{0.25cm}

\subsection{Recursive Predictions} 
We highlighted our methodology in \Cref{sec:recursive_pred_method} to carry out recursive predictions. In the following subsections, we will discuss our results of recursive prediction using the pre-trained Kinetic-Mamba models discussed above. The recursive prediction setting is considerably more challenging than one-step prediction, as errors can accumulate, amplify, as well as propagate across successive steps.

\subsubsection{Syngas A Problem with Kinetic-Mamba Framework}

\begin{figure}[htbp]
    \centering
    \includegraphics[width=\linewidth]{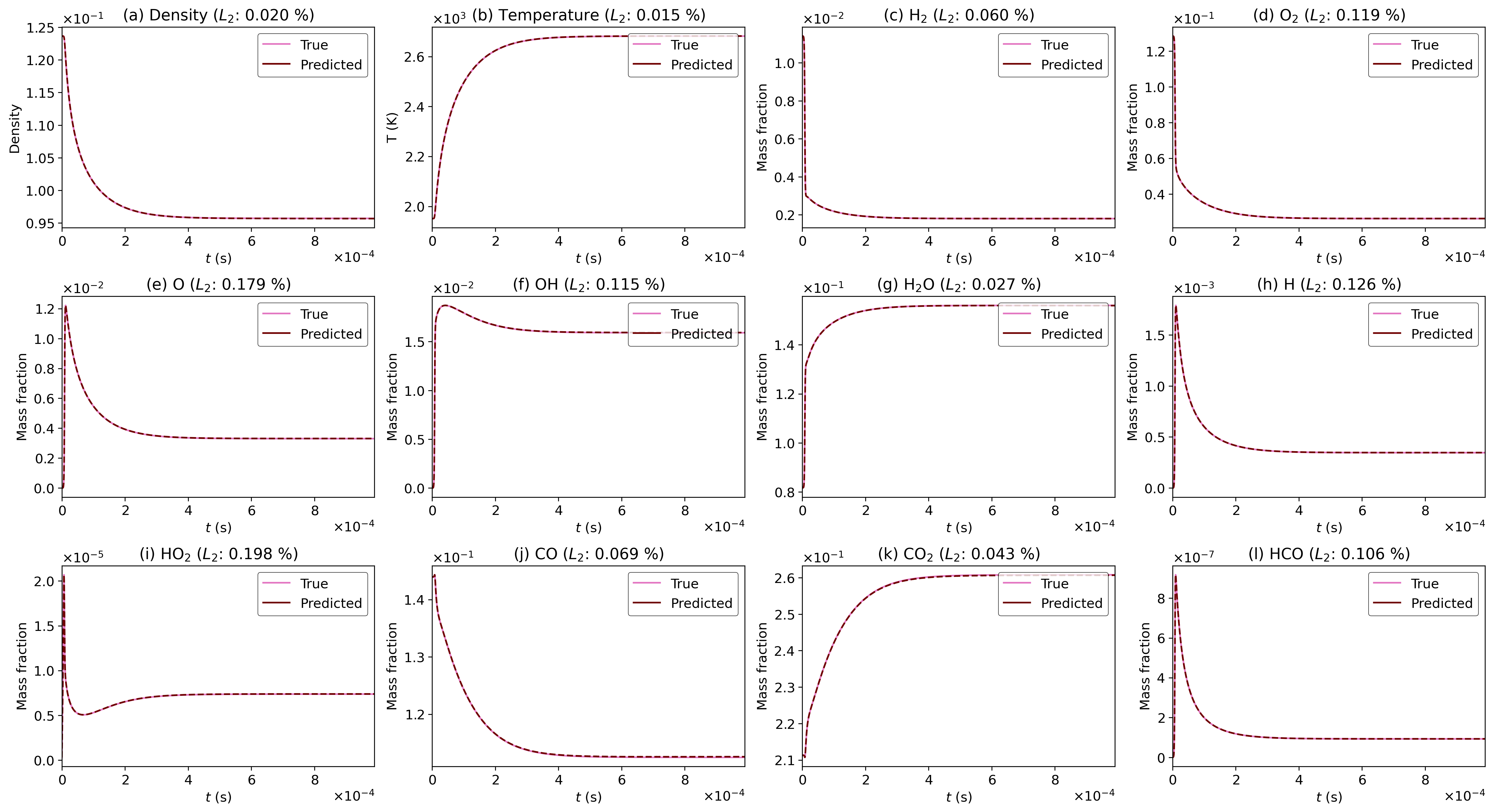}
    \caption{\textbf{Recursive prediction using KM model on Syngas A}: We used pre-trained KM model to perform recursive predictions over 99 windows of length 101.}
    \label{fig:plot_recu_SyngasA}
\end{figure}

For the recursive predictions, we use our pre-trained standalone Mamba model as described above with same hyper-parameters and parameters. We then load this model to perform recursive predictions on our test dataset. For this, we proceed to predict recursively in windows of length 101, i.e., we start with initial conditions at $t_0$ and predict until $t_{100}$. We use the predicted state at $t_{100}$ as input to the next window via the inverse-forward transform described in \Cref{sec:recursive_pred_method}, repeating for all $99$ windows. The concatenated trajectory spans $9,901$ unique time points after removing boundary duplicates. We also extract the first $9,901$ time points from the test dataset to compute the error. We decode our output back to the physical space. For every sample of each state variable within the test dataset, we tabulate the error values in \Cref{tab:errortable_rec_SyngasA} to show the relative $L_2$ error across three different runs. Over these three independent runs, we obtain a relative $L_2$ error of ($1.898 \pm 0.481$)\%. We plot the true test data corresponding to an arbitrary sample and its recursive prediction in \Cref{fig:plot_recu_SyngasA}.

\begin{table*}[htbp]
\centering
\caption{\textbf{Syngas A: Recursive Prediction}: Mean and standard deviation of the percentage error values over three runs for recursive predictions obtained using pre-trained Kinetic-Mamba model.}
\label{tab:errortable_rec_SyngasA}
\renewcommand{\arraystretch}{1.25}
\setlength{\tabcolsep}{1.25pt}
  \begin{tabularx}{\textwidth}{@{}Y
      S[table-format=1.3e+2]
      S[table-format=1.3e+2]@{}}
    \rowcolor{R111G0B0}
\thead{State Variables} & \thead{$\mu$} & \thead{$\sigma$} \\ 
\rowcolor{pink!45}\midrule
Density & 1.102e-01 & 5.100e-03 \\
Temperature & 1.169e-01 & 8.379e-03 \\
\rowcolor{pink!45}
H$_2$ & 2.891e-01 & 1.285e-02 \\
O$_2$ & 2.479e-01 & 1.037e-02 \\
\rowcolor{pink!45}
O & 6.538e+00 & 2.206e+00 \\
OH & 9.098e+00 & 2.172e+00 \\
\rowcolor{pink!45}
H$_2$O & 2.047e-01 & 2.161e-02 \\
H & 4.350e+00 & 1.591e+00 \\
\rowcolor{pink!45}
HO$_2$ & 1.052e+00 & 6.315e-02 \\
CO & 1.621e-01 & 5.943e-03 \\
\rowcolor{pink!45}
CO$_2$ & 1.280e-01 & 8.721e-03 \\
HCO & 2.358e+00 & 4.326e-01 \\
\rowcolor{pink!45}
N$_2$ & 1.411e-02 & 1.544e-04 \\ \bottomrule
\end{tabularx}
\end{table*}

\subsubsection{GRI-Mech 3.0 Problem with Latent Kinetic-Mamba Framework}
\begin{figure}[htbp]
    \centering
    \includegraphics[width=\linewidth]{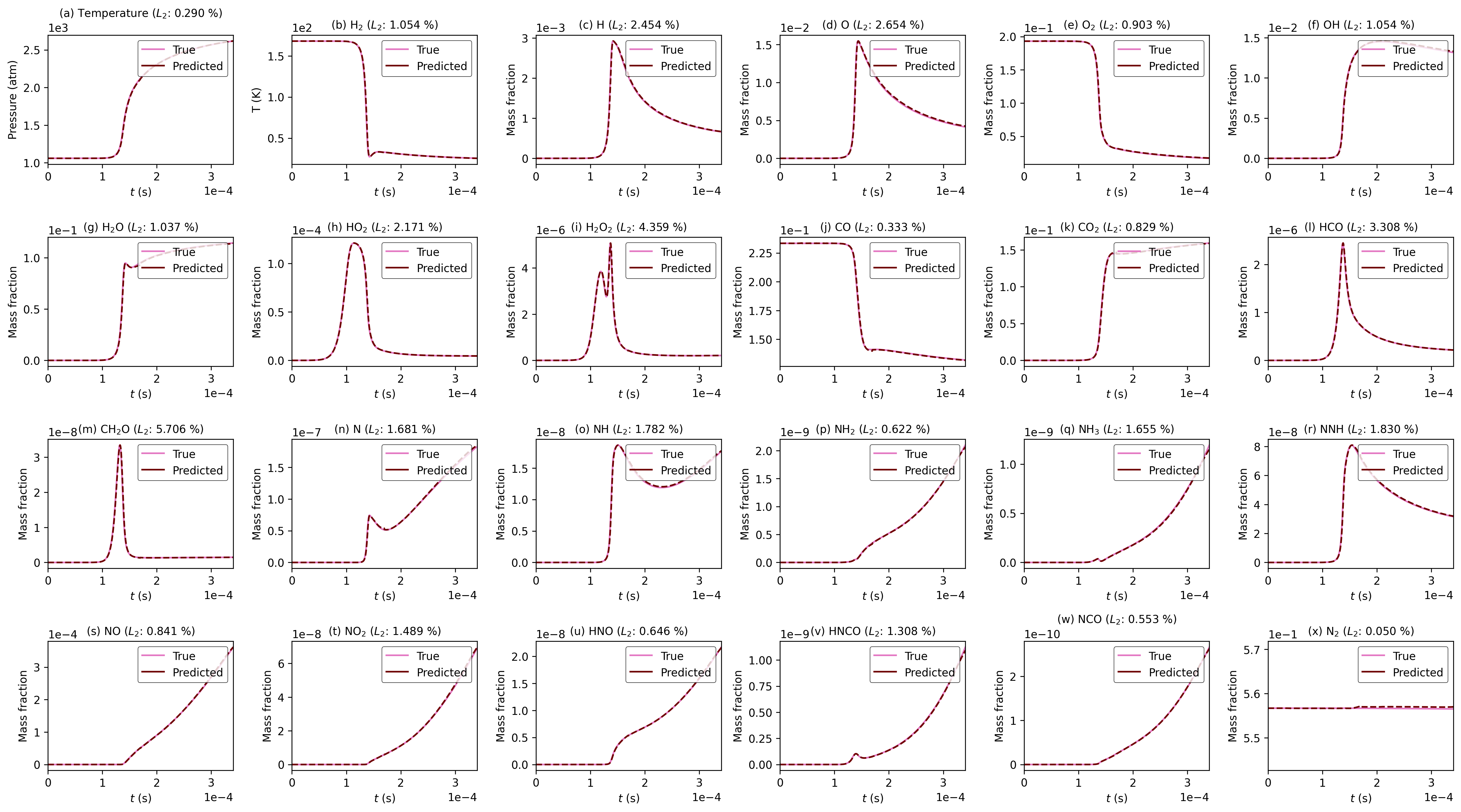}
    \caption{\textbf{Adaptive time step prediction using latent KM model on GRI-Mech 3.0}: We used latent KM model to make recursive predictions at intervals of different lengths using last time step of previous interval as input to next time step.}
    \label{fig:adaptive_step_recu_GRI}
\end{figure}

A key advantage of the Kinetic-Mamba model is its ability to perform adaptive temporal evolution during prediction. Rather than requiring a fixed window of 101 time points at each recursive step, the model is able to yield predictions over segments of variable temporal lengths as detailed in \Cref{sec:recursive_pred_method}. For instance, given CFD time steps of 10, 7.5, 3.0, \dots microseconds, one can predict over 101 time points, feed the final prediction as input, then predict the next 76 time points, and continue with a segment of length 31, and so on. \Cref{fig:adaptive_step_recu_GRI} illustrates this approach for an arbitrary sample, comparing true and predicted values across variable-length temporal segments, spanning a total of 3,405 time points. Over three independent predictions using three pre-trained Latent KM models, we obtain a mean relative error of (4.51$\pm$1.12)\%.

\subsubsection{Directions for Improvement}
The KM models efficiently predict the dynamics of the state variables in time decomposition, extrapolation and recursion. However, addressing the following can further reduce the relative error on recursive predictions. 

\par The inverse-forward transform applied at each window boundary point introduces small numerical errors that may get amplified over subsequent roll-outs. We believe that training in double precision can help substantially reduce these error values when inference is performed recursively. Additionally, augmenting the feature dimension with explicit time values, similar to the input construction for Fourier Neural Operators (FNOs) \cite{Li_2020_FNO}, can improve accuracy. However, this also introduces a possibility that the model could suffer or fail to demonstrate any drastic improvement in the accuracy, owing to the model not being able to accurately connect the temporal values with predictions within time windows of length 101, causing it to either learn spurious patterns or overfit to the dynamics of the local window on which it was trained upon. Strategies such as randomizing the time window offsets during model training can help tackle this. Moreover, training the KM models using adaptive loss functions especially designed to account for accuracy at each roll-out, or a better training methodology for including time as an additional feature can help in improving recursive predictions obtained using a pre-trained model (trained in a time decomposition setting).

\par Other directions, such as using robust error metrics that consider that relative $L_2$ errors can be dominated by near-zero denominators; understanding the means to augment normalized global or local temporal information or adopting a different methodology when performing the roll-outs are left for future work. 

\section{Conclusion\label{sec:conclusion}}
\addvspace{10pt}
In this paper, we propose a framework called Kinetic-Mamba. The Kinetic-Mamba framework is capable of accurately predicting the dynamics of stiff chemical systems using only the initial conditions as input. It also takes varied ignition regimes within the dynamics of temperature into consideration by employing different Mamba models to different regimes within the dataset. We also implemented a latent Kinetic-Mamba model, which uses principal component analysis to reduce the dimension of the state variables by converting them into latent variables, and then evolves them in time while also reconstructing the predictions back to the original state variables. We demonstrated these capabilities of the framework through Syngas and GRI-Mech 3.0 datasets where we performed training in time decomposition fashion and obtained predictions in both time-decomposed manner as well as recursively. We also observed good agreement between the predicted and the true values in extrapolation using wide range of out-of-distribution datasets that were generated using parameters that lay outside our training range. Moreover, our latent Kinetic-Mamba model performed well when trained on 99 segments of length 101 but made to recursively predict on time segments of different lengths. Through these applications of Kinetic-Mamba model, we demonstrated the accuracy of selective state space Mamba model in accurately predicting the chemistry of stiff systems.

We note that while we obtained great accuracy, the training time required by Kinetic-Mamba is much higher. In an ideal case of integration of the framework with CFD solver, this cost is amortized by the fast prediction times, since training is mostly done offline. We list the training and prediction time in \Cref{Appendix:CompTime} for our different demonstrative examples. We also note that higher training times can be attributed to the higher number of samples that the model is trained upon as a result of time decomposition method. Additionally, since the main focus of this paper is to illustrate the application of our Mamba-based Kinetic-Mamba model when applied to stiff chemical kinetics, the training was done in single precision.

As a part of future work, we want to focus on the extension of the Kinetic-Mamba framework to support double precision data resolution during network training. We also want to focus on model training strategies that can enhance the accuracy during recursive prediction. Kinetic-Mamba's performance on recursive predictions can be ascribed to its failure to be informed about the stiffness in the dynamics and when they might occur, as well as to the skewed information that state space model in general receives due to localization of dynamics within the prediction windows. We discussed several pathways to improve accuracy in recursive predictions, such as, augmenting the feature dimension with additional information like time values, performing training with a different methodology using adaptive loss functions, specifically designed to increase accuracy on recursive predictions, as well as by training the model in double precision. We believe that these pathways can lead to interesting future directions. Future works can also focus on training parallelization strategies that can help reduce the runtime when training subsets from a bigger dataset in the presence of varied ignition regimes. 

To summarize, in this study, we presented a Kinetic-Mamba framework that helps in accurately predicting the dynamics of the thermochemical state variables, via its standalone Mamba model, ignition regime-informed Mamba models, mass-conserving Mamba model and latent Mamba model, all trained in time-decomposition fashion. The Kinetic-Mamba framework is adept at time decomposed predictions as well as in extrapolation tasks, and shows reasonable accuracy in predictions obtained recursively. Through this work, we demonstrate that Mamba-based models provide a scalable alternative to implicit solvers for stiff multiscale dynamical systems. This illustrates the ability of selective state space models emerging as a cornerstone in future studies based on combustion problems involving chemical kinetics of complex reactions.

\section*{Acknowledgment}
 The authors acknowledge the support from the Small Business Technology Transfer (STTR) program, USA (Grant No: GR5291245) and Defense Threat Reduction Agency (DTRA), USA. This research was conducted using computational resources available at the Center for Computation and Visualization, Brown University. The implementation of standalone Mamba models is inspired by \cite{Hu_2024_state} and the base code of Mamba models from \cite{Gu_2023_mamba}. 

\section*{Data Availability Statement}
Data will be made available upon request to the corresponding author.

\section*{AI Disclosure Statement}
During the preparation of this work, the authors used Claude (Anthropic) to enhance the readability of the manuscript. After using the tool, the authors reviewed and edited the content as needed, and take full responsibility for the content of the publication.

\section*{Declaration of competing interests} 
The authors declare that they have no known competing financial interests or personal relationships that could have appeared to influence the work reported in this paper.

% -------------------------------------------------------------------- %
% -------------------------------------------------------------------- %
% -------------------------------------------------------------------- %

 \baselineskip 9pt

% -------------------------------------------------------------------- %
% -------------------------------------------------------------------- %
% -------------------------------------------------------------------- %

\bibliographystyle{unsrtnat} % or plainnat, abbrvnat, etc.
\bibliography{references}

\begin{appendices}
In the following sections, we shall touch upon several relevant aspects corresponding to the data, implementation and results that we could not include in the main body of the paper.
\section{Range of Data \label{Appendix:DataRange}}
In this section, we report the maximum and minimum values from the training and testing dataset. For Syngas A dataset, the maximum and minimum values for both training and testing dataset is tabulated in \Cref{tab:SyngasA_maxmin}, and for Syngas B we divided the dataset into two parts, as mentioned above, and we tabulate the training and testing dataset minimum and maximum values in \Cref{tab:SyngasB_bif_minmax}. We also have \Cref{tab:GRI_minmax} tabulating the maximum and minimum values from the GRI dataset.
\vspace{0.25cm}
\begin{table*}[htbp]
  \centering
  \renewcommand{\arraystretch}{1.25}
  \setlength{\tabcolsep}{1.25pt}
  \caption{Syngas A dataset values which is generated with equivalence ratios and temperatures from a uniform distribution.}
  \begin{tabularx}{\textwidth}{@{}Y
      S[table-format=1.3e+2]
      S[table-format=1.3e+2]@{}}
    \rowcolor{R111G0B0}
    \thead{State Variables} & \thead{Max} & \thead{Min} \\
    \midrule
    \rowcolor{pink!45}
    Density & 7.062e-01 & 5.239e-02 \\
    Temperature & 2.984e+03 & 1.000e+03 \\ \rowcolor{pink!45}
    H$_2$ & 2.926e-02 & 8.978e-06 \\
    O$_2$ & 2.153e-01 & 6.199e-05 \\ \rowcolor{pink!45}
    O & 2.601e-02 & 1.298e-12 \\
    OH & 2.848e-02 & 8.181e-13 \\ \rowcolor{pink!45}
    H$_2$O & 2.497e-01 & 1.754e-03 \\
    H & 5.538e-03 & 7.335e-13 \\ \rowcolor{pink!45}
    HO$_2$ & 3.289e-04 & 1.727e-12 \\
    CO & 3.270e-01 & 6.264e-04 \\ \rowcolor{pink!45}
    CO$_2$ & 4.839e-01 & 2.134e-05 \\
    HCO & 5.301e-06 & 0.000e+00 \\ \rowcolor{pink!45}
    N$_2$ & 7.091e-01 & 3.260e-01 \\
    \bottomrule
  \end{tabularx}
  \label{tab:SyngasA_maxmin}
\end{table*}

\begin{table*}[htbp]
  \centering
  \renewcommand{\arraystretch}{1.25}
  \caption{Syngas B dataset values generated with equivalence ratios and temperatures sampled from a random distribution.}
  \setlength{\tabcolsep}{1.25pt}
  \begin{tabularx}{\textwidth}{@{}Y
      S[table-format=1.3e+2]
      S[table-format=1.3e+2]@{}}
    \rowcolor{R111G0B0}
    \thead{State Variables} & \thead{Max} & \thead{Min} \\
    \midrule
    \rowcolor{pink!45}
Pressure & 2.606e+00 & 1.010e+00 \\
Temperature & 3.164e+03 & 5.006e+02 \\
\rowcolor{pink!45}
H$_2$ & 4.009e-03 & 7.654e-05 \\
O$_2$ & 1.825e-01 & 1.365e-02 \\
\rowcolor{pink!45}
O & 3.202e-02 & 2.402e-24 \\
OH & 1.545e-02 & 7.039e-27 \\
\rowcolor{pink!45}
H$_2$O & 2.870e-02 & 4.609e-24 \\
H & 1.424e-03 & 5.883e-28 \\
\rowcolor{pink!45}
HO$_2$ & 1.009e-04 & 1.106e-22 \\
CO & 2.785e-01 & 3.390e-02 \\
\rowcolor{pink!45}
CO$_2$ & 2.555e-01 & 8.990e-22 \\
HCO & 2.297e-06 & 5.353e-30 \\
\rowcolor{pink!45}
N$_2$ & 6.284e-01 & 5.811e-01 \\
    \bottomrule
  \end{tabularx}
  \label{tab:SyngasB_bif_minmax}
\end{table*}

\begin{table*}[htbp]
  \centering
  \renewcommand{\arraystretch}{1.25}
  \caption{GRI-Mech 3.0 dataset values generated with equivalence ratios and temperatures sampled from a uniform distribution.}
  \setlength{\tabcolsep}{1.25pt}
  \begin{tabularx}{\textwidth}{@{}Y
      S[table-format=1.3e+2]
      S[table-format=1.3e+2]@{}}
    \rowcolor{R111G0B0}
    \thead{State Variables} & \thead{Max} & \thead{Min}\\
    \midrule
    \rowcolor{pink!45}
    Temperature & 2.870e+03 & 1.000e+03 \\
    H$_2$ & 1.787e-02 & 5.446e-05 \\ \rowcolor{pink!45}
    H & 3.873e-03 & 5.110e-14 \\
    O & 2.356e-02 & 1.160e-12 \\ \rowcolor{pink!45}
    O$_2$ & 2.298e-01 & 6.809e-03 \\
    OH & 1.921e-02 & 3.626e-14 \\ \rowcolor{pink!45}
    H$_2$O & 1.250e-01 & 1.104e-14 \\ 
    HO$_2$ & 1.428e-04 & 1.681e-12 \\ \rowcolor{pink!45}
    H$_2$O$_2$ & 9.155e-06 & 1.595e-17 \\
    CO & 2.482e-01 & 4.318e-03 \\ \rowcolor{pink!45}
    CO$_2$ & 1.871e-01 & 3.257e-12 \\
    HCO & 2.748e-06 & 5.465e-17 \\ \rowcolor{pink!45}
    CH$_2$O & 4.200e-08 & 9.630e-21 \\
    N & 2.051e-06 & 2.698e-27 \\ \rowcolor{pink!45}
    NH & 2.753e-07 & 1.500e-28 \\
    NH$_2$ & 5.226e-08 & 3.043e-31 \\ \rowcolor{pink!45}
    NH$_3$ & 3.004e-08 & 5.108e-33 \\
    NNH & 1.079e-07 & 5.047e-19 \\ \rowcolor{pink!45}
    NO & 1.280e-02 & 7.873e-27 \\
    NO$_2$ & 6.751e-06 & 5.099e-37 \\ \rowcolor{pink!45}
    HNO & 6.917e-07 & 2.777e-30 \\
    HNCO & 2.934e-08 & 2.370e-31 \\ \rowcolor{pink!45}
    NCO & 6.952e-09 & 5.533e-31 \\
    N$_2$ & 6.623e-01 & 5.411e-01 \\
    \bottomrule
  \end{tabularx}
  \label{tab:GRI_minmax}
\end{table*}

\section{Computational Time \label{Appendix:CompTime}}
In this section, we shall list the amount of time it took to train the Kinetic-Mamba framework on different problems using their respective training dataset that we discussed in this paper, as well as the amount of time it took to make a prediction on the test dataset. We want to mention that the models were trained on an NVIDIA L40S GPU (\url{https://www.nvidia.com/en-us/data-center/l40s/}) in a network cluster. 
\begin{enumerate}
    \item In the problem corresponding to the Syngas A dataset \Cref{subsec: Problem1}, we observe that the total training time is 161,315.47 seconds (44.81 hours) and the prediction time is 2.77 seconds. 
    \item When we transition to the mass-conserving Kinetic-Mamba framework on the Syngas A dataset \Cref{subsec: Problem2}, we observe that the total training time is 161,677.78 seconds (44.91 hours) and the prediction time is 2.74 seconds. Moreover, it takes 142.72 seconds to go from the 
    $m$-dimensional manifold to the
    $m+1$-dimensional manifold.
    \item When we use the Syngas B dataset with our Kinetic-Mamba framework which takes the ignition regimes into consideration \Cref{subsec: Problem3}, we note that the training time is 12,743.51 seconds (3.54 hours) and 185,055.27 seconds (51.40 hours) corresponding to the data lying below $\tau$K and above $\tau$K and the prediction times are 0.78 seconds and 4.38 seconds, respectively.
    \item When considering the latent Kinetic-Mamba model on the GRI dataset \Cref{subsec:GRIProblem}, the training time is 45,764.18 seconds (12.71 hours) and the prediction time is 0.92 seconds.
\end{enumerate}

\section{Additional Results on Extrapolation \label{appendix:additional_recursion_syngasA}}
We mentioned inflation in the relative $L_2$ error observed in violin plot \Cref{fig:extrapolation} due to small value of the denominator (true value) due to the magnitudes of mass fractions of species. Hence, we clip the denominator to $\max(den,\epsilon)$, where $\epsilon = 10^{-3} \times \tau$ where $\tau$ is the sample mean of the relative $L_2$ error of true values with respect to time.

\begin{figure}
    \centering
    \includegraphics[width=\linewidth]{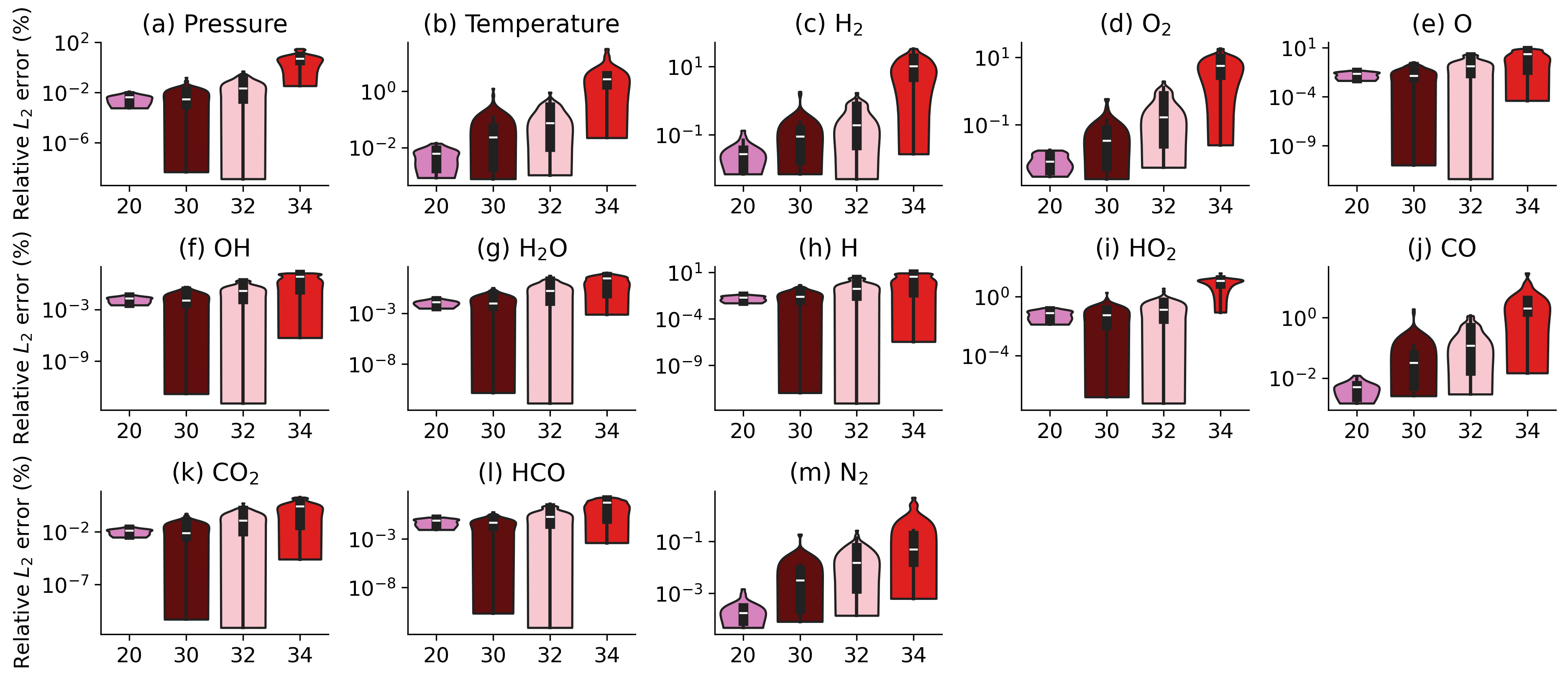}
    \caption{\textbf{Syngas B violin plot using revised error metric during extrapolation}: Relative $L_2$ error for Syngas B extrapolation dataset when extrapolation was performed using pre-trained model with error values depicted in log scale on y-axis using the clipped denominator error metric.}
    \label{fig:placeholder}
\end{figure}

\end{appendices}

% -------------------------------------------------------------------- %
% -------------------------------------------------------------------- %
% -------------------------------------------------------------------- %

\baselineskip 10pt

% -------------------------------------------------------------------- %
% -------------------------------------------------------------------- %
% -------------------------------------------------------------------- %

% -------------------------------------------------------------------- %
% -------------------------------------------------------------------- %
% -------------------------------------------------------------------- %

\end{document}